%% file: girnet_arxiv.tex
\documentclass[10pt,twocolumn,letterpaper]{article}

\usepackage{multirow,booktabs}
\usepackage[pagenumbers]{cvpr}
\input{preamble}

\usepackage{adjustbox}

\definecolor{cvprblue}{rgb}{0.21,0.49,0.74}
\usepackage[pagebackref,breaklinks,colorlinks,citecolor=cvprblue]{hyperref}

\def\paperID{1580} 
\def\confName{CVPR}
\def\confYear{2024}

\title{Guided Image Restoration via Simultaneous Feature and Image Guided Fusion}

\author{Xinyi Liu$^{1,3}$, Qian Zhao$^{1}$, Jie Liang$^{2,3}$, Hui Zeng$^{3}$, Deyu Meng$^{1}$, Lei Zhang$^{2,3}$\\
\hspace{-0.2cm}$^{1}$Xi'an Jiaotong University, \quad
\hspace{-0.2cm}$^{2}$The HongKong Polytechnic University, \quad 
\hspace{-0.2cm}$^{3}$OPPO Research }

\begin{document}
\maketitle
\input{sec/0_abstract}    
\input{sec/1_intro}

\input{sec/2_relatedwork}
\input{sec/3_girnet}
\input{sec/4_exp}
\input{sec/5_conclusion}

\clearpage
{
    \small
    \bibliographystyle{ieeenat_fullname}
    \bibliography{main}
}

\input{sec/X_suppl}

\end{document}

%% file: preamble.tex
\usepackage[dvipsnames]{xcolor}


%% file: sec/0_abstract.tex
\begin{abstract}
Guided image restoration (GIR), such as guided depth map super-resolution and pan-sharpening, aims to enhance a target image using guidance information from another image of the same scene. Currently, joint image filtering-inspired deep learning-based methods represent the state-of-the-art for GIR tasks. Those methods either deal with GIR in an end-to-end way by elaborately designing filtering-oriented deep neural network (DNN) modules, focusing on the feature-level fusion of inputs; or explicitly making use of the traditional joint filtering mechanism by parameterizing filtering coefficients with DNNs, working on image-level fusion. The former ones are good at recovering contextual information but tend to lose fine-grained details, while the latter ones can better retain textual information but might lead to content distortions. In this work, to inherit the advantages of both methodologies while mitigating their limitations, we proposed a Simultaneous Feature and Image Guided Fusion (\textbf{SFIGF}) network, that simultaneously considers feature and image-level guided fusion following the guided filter (GF) mechanism. In the feature domain, we connect the cross-attention (CA) with GF, and propose a GF-inspired CA module for better feature-level fusion; in the image domain, we fully explore the GF mechanism and design GF-like structure for better image-level fusion. Since guided fusion is implemented in both feature and image domains, the proposed SFIGF is expected to faithfully reconstruct both contextual and textual information from sources and thus lead to better GIR results. We apply SFIGF to 4 typical GIR tasks, and experimental results on these tasks demonstrate its effectiveness and general availability.

\end{abstract}

%% file: sec/1_intro.tex
\section{Introduction}
With the rapid development of photography techniques, images can be captured from multiple sensors simultaneously under the same scene. Consequently, we can expect higher quality imaging by making use of rich and complementary information delivered by those multi-modal source images. 
Considering the case with two source images, one may be interested in restoring one of these two images with the guide information transferred from the other one \cite{he2010guided, he2012guided}, which can be referred to as guided image restoration (GIR), including guided denoising \cite{xiong2021mffnet}, guided depth map super-resolution (GDSR) \cite{hui2016depth}, pan-sharpening \cite{yang2017pannet}, and guided low-light raw image enhancement (LRIE) \cite{dong2022abandoning}.

Traditionally, the GIR is mainly realized by image filters, including bilateral filter (BF) \cite{tomasi1998bilateral}, guided filter (GF) \cite{he2010guided}, and their variants \cite{he2012guided, xiao2012fast}.
Such joint filters can exploit complementary information across sources, and transfer the extracted textual contents from the guide image to the target. 
However, they highly rely on the manually designed filtering mechanism, which may not be flexible enough to deal with complex image structures in real applications.

In the last decade, motivated by their successful applications, deep learning methods have been introduced to the GIR tasks. A straightforward way is to directly feed the input source images to a deep neural network (DNN), mostly with filtering-inspired modules, and output the restoration result in an end-to-end way \cite{li2016djif,deng2020cunet,kim2021deformable}. Observing that well-designed DNNs are effective feature extractors, such a strategy indeed realizes image fusion mainly in the feature domain. Another way is to explicitly take the joint filtering mechanism into consideration, and learn components of filters by DNNs \cite{wu2018fgf, pan2019spatially}. Such methods directly perform GIR at the image level, which inherits the advantage of joint filtering, while being more flexible. Thanks to the powerful learning ability, both types of deep learning methods can perform well on a specified GIR task after proper training.

\begin{figure}[t]
	\centering
        \includegraphics[width=0.95\linewidth]{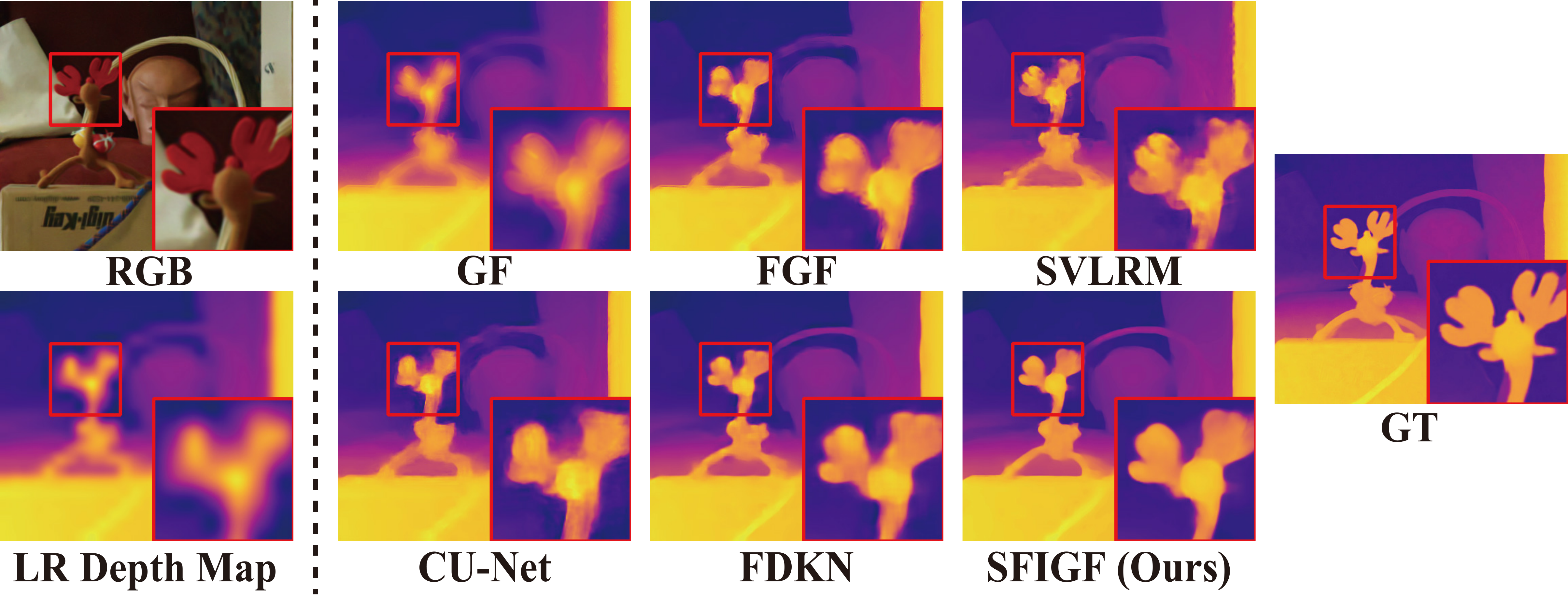}
        \vspace{-0.3cm}
	\caption{Visualization of current filtering-inspired GIR methods.}
	\vspace{-0.6cm}
	\label{fig:intro} 
\end{figure}

However, these deep learning approaches still have limitations.
In specific, the former strategy which mainly implements feature-level fusion, though good at restoring the contextual information of images, e.g., removing noise, may lose fine-grained textual information, e.g., edges, delivered by the guide image.
In contrast, the latter methodology can do better in transferring the textual information from the guidance to the target at the image level, but tends to result in more unexpected distortions of image contents.
These can be intuitively observed by the GDSR example shown in Fig.~\ref{fig:intro}.
It can be seen that the results of CU-Net \cite{deng2020cunet} and  FDKN \cite{kim2021deformable}, which belong to the former type methods, look smoother, but the edges are somewhat blurry. In comparison, FGF \cite{wu2018fgf} and SVLRM \cite{pan2019spatially} transfer more textual information from the guide RGB image, but lead to unnatural distortions in contents, such as ghost effects. These observations motivate us to leverage the benefits of both methodologies while mitigating their limitations.
In this work, we propose a new DNN framework, called  Simultaneous Feature and Image Guided Fusion (SFIGF), for the GIR tasks.
The proposed SFIGF combines the advantages of both of the aforementioned methodologies in using DNNs, by considering guided fusion operations simultaneously in feature and image domains. 
Specifically, in the feature domain, by exploring the connections between cross-attention (CA) \cite{vaswani2017attention, tan2019lxmert} and guided filter (GF) \cite{he2010guided}, we propose a GF-inspired cross-attention (GICA) module, which integrates fusion mechanisms of GF with efficient structures of transformer; 
in the image domain, SFIGF explicitly fuses input source images in the guided filtering fashion with key parameters predicted by specifically designed network modules, sharing the similar idea of FGF \cite{wu2018fgf} and SVLRM \cite{pan2019spatially}.
The guided fusion outputs from the image and feature domains are then aggregated for the final GIR result. Since the final result is contributed by both feature and image domains, it is expected to recover as much as possible textual information from the guide image, while also faithfully restoring the contextual information from source images.

Our contributions can be summarized as follows:
\begin{itemize}
\item By exploring the connection between CA and GF, we propose the GICA module, which enhances the CA with the fusion principles of GF. Based on the GICA block, we design a feature-guided fusion (FeGF) module to fuse multi-scale features extracted from source images. The FeGF module can not only better capture the spatial dependencies due to the CA operation, but also better fuse the complementary information from sources by virtue of the guided filtering mechanism.

\item We design an image-guided fusion (ImGF) module for image-level fusion, which inherits the idea of learnable GF \cite{wu2018fgf, pan2019spatially}, but more coincides with GF; and also a cooperative multi-scale feature extraction (CMFE) module based on the NAF blocks \cite{chen2022nafnet} for cooperatively extracting features from source images. Together with the FeGF module, we construct the SFIGF for GIR tasks, which is expected to leverage the advantages of both methodologies discussed before.

\item  We apply the proposed SFIGF to 4 typical GIR tasks, including GDSR, pan-sharpening, multi-frame image fusion (MFIF), and guided LRIE.
Extensive experiments on these tasks validate the effectiveness of SFIGF and demonstrate its wide applicability.
\end{itemize}

%% file: sec/2_relatedwork.tex
\section{Related Work}
{\bf Traditional GIR methods.} Traditionally, GIR tasks are mainly addressed by joint image filtering, where the filtering is implemented on the to-be-restored image, while the coefficients of the filter are computed by also considering the guidance image. Many joint image filters were designed to achieve this goal, such as BF \cite{tomasi1998bilateral}, GF \cite{he2010guided, he2012guided} and their variants \cite{li2014weighted, xiao2012fast}.
Another type of methodology to deal with GIR is to formulate an optimization model for a certain problem, such as Markov random field labeling for GDSR \cite{mac2012patch} and variational optimization for pan-sharpening \cite{wang2018bayesian}.
Comparing the two methodologies, the latter is more task-specific, while the former is more general and has wider applications \cite{he2012guided, li2014weighted, tomasi1998bilateral}.
However, both methodologies highly rely on manually designing according to subjective priors of tasks, and thus the performance is often limited.

\noindent{\bf Deep learning-based GIR methods.} Motivated by their success in other applications, especially in image restoration tasks, deep learning methods have dominated the GIR tasks in recent years.

Since joint image filtering has shown its effectiveness and taken an important position in traditional GIR methods, researchers have attempted to improve it by using deep learning methodology. 
For example, Wu et al.\cite{wu2018fgf} proposed to replace average operators with convolution layers for a learnable GF; Pan et al. \cite{pan2019spatially} took a step further by learning GF coefficients parameterized by neural networks. Such methods intrinsically realize GIR at the image level, which is good at transferring the textual information from the guide image to the target, but the performance is limited by the explicit joint filtering mechanism.

Other than explicitly using filtering mechanisms in the image domain, more researchers have tried to design end-to-end DNNs by introducing filtering-oriented structures. For example, \cite{li2016djif, li2019joint, deng2020cunet} proposed to use convolution-based sub-networks to split common and salient features from sources for final fusion results, Kim et al. \cite{kim2021deformable} built filtering-inspired networks with deformable convolution for adaptive filtering location, and Zhong et al. \cite{zhong2021dagif} and Su et al. \cite{Su_2019_pac} tried to predict kernels with learnable weights. Beyond the filtering-inspired ones, other techniques have also been considered, such as algorithm unrolling \cite{deng2019deep, madunet}, densely connection \cite{xu2020fusiondn} and transformer \cite{zhao2023cddfuse}. As mentioned in the Introduction, these methods, though good at contextual recovery, tend to lose fine-grained textual information of images.

%% file: sec/3_girnet.tex
\section{Preliminary and Motivation}\label{sec:pre}
Before presenting our SFIGF, we need to first review GF and CA, and then explore the connection between them to motivate our work.

\subsection{Guided filter}\label{sec:gf}
GF \cite{he2010guided, he2012guided} is a versatile image processing tool, originally designed for filtering an input image with a guidance image.
Due to its ability in effectively fusing two source images by making use of complementary information delivered by them, GF has been applied to a variety of image processing tasks, including image fusion \cite{ShutaoLiImageFusion2013, bavirisetti2017medicalfusion} and segmentation \cite{zhang2022segmentation}.
The key assumption under GF is that the filtering output $Q$ is a linear transform of the guidance image $I$ in a window $\omega_{k}$ centered at the pixel $k$:
\vspace{-2mm}
\begin{equation}\label{eq:1}
	Q_{j} = a_{k}I_{j} + b_{k},~~~ \forall j\in\omega_{k},
	\vspace{-2mm}
\end{equation}
where $a_{k}$ and $b_{k}$ are the 
coefficients in the window $\omega_{k}$, and the subscript $j$ refers to the $j$th pixel in $\omega_k$. The coefficients can be obtained by minimizing the following cost function:\vspace{-2mm} 
\begin{equation}\label{eq:2}
	E(a_k, b_k) = \sum\nolimits_{j \in \omega_{k}}{\left((a_k I_j + b_k - P_j)^2 + \epsilon a_k^2\right)},
	\vspace{-2mm}
\end{equation}
which ensures that the output $Q$ should also be close to input $P$. Here, $\epsilon$ is a regularization parameter preventing $a_{k}$ from being too large. The solution can be analytically computed:\vspace{-2.2mm}
\begin{gather}
	a_k = \frac{\frac{1}{|\omega|}\sum_{j\in\omega_{k}}{I_j P_j-\mu_k\bar{P}_k}}{\sigma_k^2 + \epsilon} = \frac{{\mathrm{Cov}(I,P)}_{k}}{\sigma_k^2 + \epsilon}, \label{eq:a_gf}\\
	b_k = \bar{P}_k -a_k\mu_k.\label{eq:b_gf}
\end{gather}
\vskip -2.2mm
\noindent Here, $\mu_k$ and $\sigma_k^2$ are the mean and variance of $I$ in $\omega_{k}$, $|\omega|$ is the number of pixels in $\omega_k$, $\bar{P}_k=\frac{1}{|\omega|}\sum_{j\in\omega_{k}}{P_j}$ is the mean of $P$ in $\omega_k$ and $\mathrm{Cov}(\cdot,\cdot)_k$ indicates the covariance. Then the filtering output can be finally calculated by:\vspace{-2mm}
\begin{equation}
	Q_j = \frac{1}{|\omega|} \sum\nolimits_{k:j\in\omega_k}{(a_k I_j+b_k)}=\bar{a}_j I_j +\bar{b}_j,
	\label{eq:6}
	\vspace{-2mm}
\end{equation}
where $\bar{a}_j = \frac{1}{|\omega|}\sum_{k\in\omega_{j}}{a_k} $ and $\bar{b}_j = \frac{1}{|\omega|}\sum_{k\in\omega_{j}}{b_k} $ by averaging all the possible values of $Q_j$ involved in all windows that contain $j$. Eq.~\eqref{eq:6} can also be written with matrix form:\vspace{-2mm}
\begin{equation}
	Q_{GF}=A_{GF}\circ I+B_{GF},
	\label{eq:GF_matrix}
	\vspace{-2mm}
\end{equation}
where $A_{GF}$ and $B_{GF}$ are composed of $\bar{a}_i$s and $\bar{b}_i$s, respectively, and $\circ$ denotes the element-wise multiplication.
It can be seen that $A_{GF}$ explicitly depicts the correlation between the inputs $I$ and $P$ due to the covariance included in Eq.~\eqref{eq:a_gf}, and thus is able to reflect the structural information in $I$ that is related to $P$. 

\subsection{Cross-attention in DNNs}\label{sec:ca}
Recently, Transformers \cite{vaswani2017attention, liang2021swinir, hassani2023neighborhood} have become dominant DNN architectures in various tasks \cite{liang2021swinir, khan2022transformers}. One of the key components of them is the attention mechanisms \cite{bahdanau2014neural}. Initially, self-attention (SA) \cite{vaswani2017attention} was used to effectively capture dependencies within individual modalities or sequences. 
To model complex interactions between multiple sources, CA \cite{tan2019lxmert} was proposed to generalize the attention mechanisms through cross-domain queries. 
Mathematically, for patch tokens $x, y \in \mathbb{R}^C$, CA mechanism can be expressed as:
\vspace{-3mm}
\begin{gather}
    Q_x=xW_q,
    K_y=yW_k,
    V_y=yW_v,\\ 
    A(x,y) = \mathrm{Softmax}\left(\frac{Q_xK_y^T}{\sqrt{d}}\right), \label{eq:a}\\
    \mathrm{CA}(x,y)= A(x,y)V_y
    =\mathrm{Softmax}\left(\frac{Q_xK_y^T}{\sqrt{d}}\right)V_y, \label{eq:cross}
\end{gather}
\vskip -3mm
\noindent where $W_q, W_k, W_v \in \mathbb{R}^{C\times d}$ are learnable parameters, $C$ is the token dimension, $d$ is the scaling parameter. In this way, CA allows querying relationships between queries ($Q_x$) from one space and keys ($K_y$) and values ($V_y$) encoded in another space, and thus can well capture the interdependencies across different sources.

\subsection{Connecting CA to GF}\label{sec:discussion}
Though looks very different at first glance, the GF and CA indeed share some similar mechanisms. Specifically, if we treat $x$ and $y$ in CA as $P$ and $I$ in GF, respectively, we can see from Eq.~\eqref{eq:cross} that the calculation of $A(x, y)$ is very similar to the calculation of $A_{GF}$, both of which are trying to capture the correlations between inputs. Besides, the $V_y$ term is indeed a transformation of $y$. Therefore, the overall process of CA, i.e., $A(x,y)V_y$ can be regarded as a kind of generalization of the $A_{GF}\circ I$ term in GF. The only intrinsic difference between GF and CA is the $B_{GF}$ term, which can in some sense be seen as the residuals from $P$, by eliminating the information correlated to $I$.
Since GF has been shown to be effective in fusing source images at the image level, while CA is powerful in feature-level information fusion, the above connection between them motivates us to integrate them together to build a GF-inspired cross-attention (GICA) block (Section \ref{sec:fegf}), seeking for a better feature-level fusion for the GIR tasks. In addition, we aim to maintain the GF mechanism at the image level, drawing upon similar ideas as previous studies \cite{pan2019spatially,wu2018fgf}, which together with the feature-level fusion establishes the architecture of the proposed SFIGF.

\section{Proposed Method}
Now we are ready to propose the SFIGF. In the following, we first introduce the overall structure and workflow of SFIGF and then provide details for each building block.
\subsection{Overall network structure and workflow}
\begin{figure}[t]
	\setlength{\abovecaptionskip}{0.2cm}
	\centering
	\includegraphics[width=0.95\linewidth]{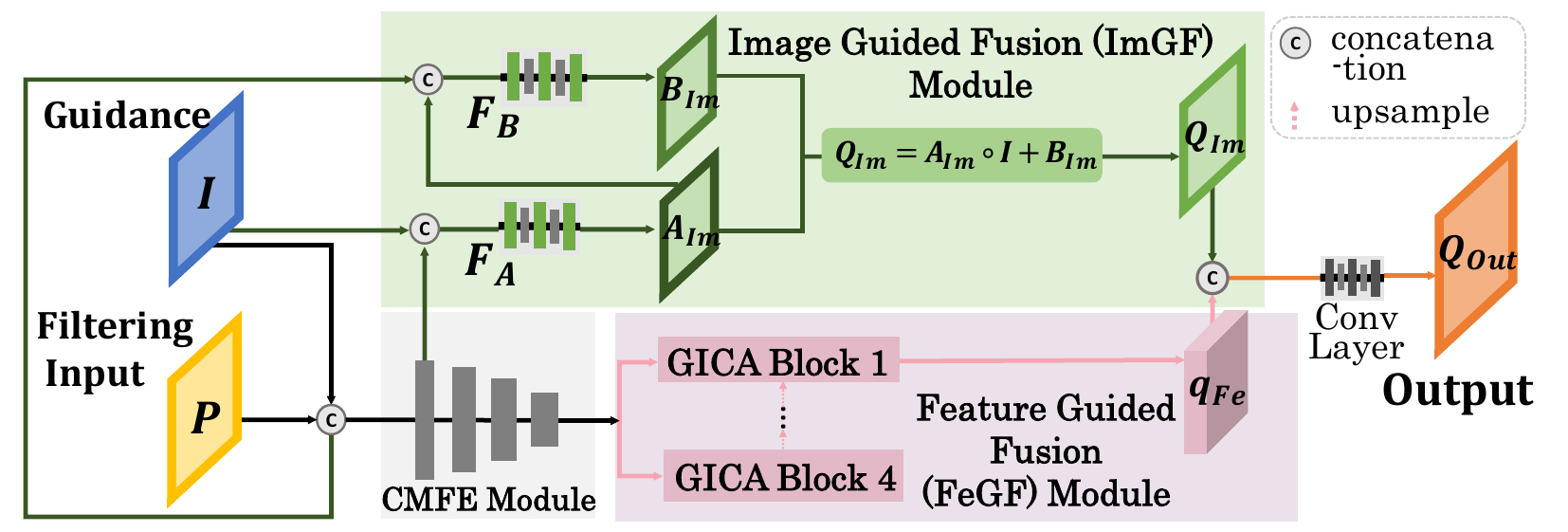}
	\vspace{-0.1cm}
	\caption{Overview of the SFIGF. The CMFE, ImGF, and FeGF modules are introduced in Section~\ref{sec:smfe}, \ref{sec:imgf}, and \ref{sec:fegf}, respectively.} 
	\label{fig:girnet}
	\vspace{-0.55cm}
\end{figure}

As shown in Fig.~\ref{fig:girnet}, our proposed SFIGF comprises three key modules: a cooperative multi-scale feature extraction (CMFE) module for extracting multi-scale features from different sources, an FeGF module for feature-level fusion, and an ImGF module for image-level fusion. In particular, the FeGF module mainly contributes to better recovering contextual information from diverse types of degradations by the proposed GICA block, while the ImGF module is mainly designed to better preserve the original details of source images by the explicit GF mechanism.  

Given the to-be-restored image $P$ and the guidance $I$ as input, SFIGF outputs the restoration $Q_{Out}$ in the following way. First, $P$ and $I$ are input to the CMFE module to jointly extract their multi-scale features. The extracted features are then passed to the ImGF and FeGF modules for subsequent processing. In the FeGF module, the extracted multi-scale features are first fused by GICA blocks at each scale and then aggregated by upsample blocks to obtain the final feature $q_{Fe}$. In the ImGF module, the extracted features, together with the source images, are used to infer $A_{Im}$ and $B_{Im}$ for the image-level fusion result $Q_{Im}$. Finally, $q_{Fe}$ and $Q_{Im}$ are concatenated and fed into a small network with attention and convolutional layers to obtain $Q_{Out}$.

\subsection{CMFE for multi-scale feature extraction}\label{sec:smfe}
\begin{figure}[t]
	\setlength{\abovecaptionskip}{0.1cm}
	\centering
	\includegraphics[width=0.95\linewidth]{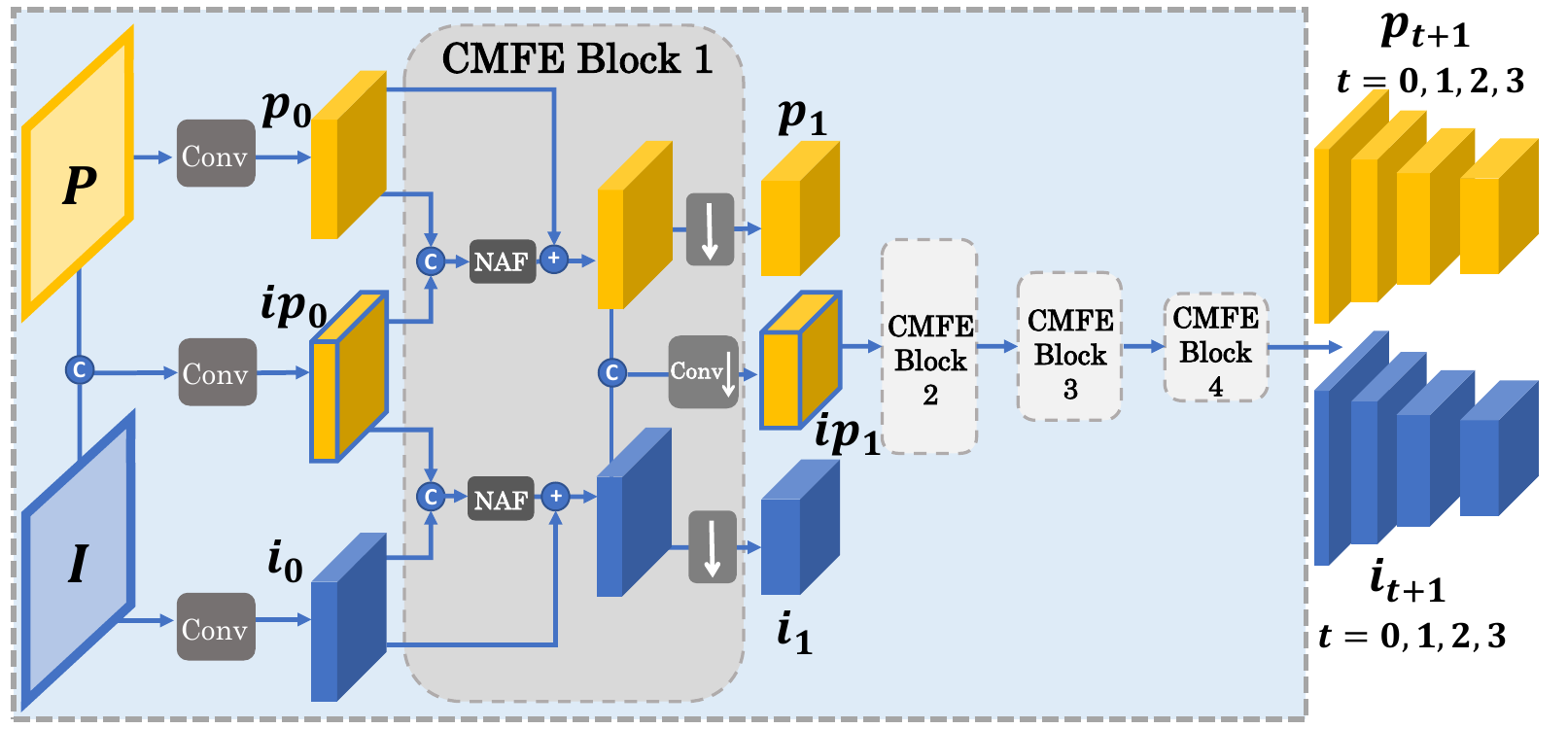}
	\caption{Overview of the CMFE module introduced in Section~\ref{sec:smfe}. 
	NAF refers to the NAF-block \cite{chen2022nafnet}.}
	\label{fig:smfe}
	\vspace{-0.55cm}
\end{figure}

SFIGF first needs to extract features from the guidance image $I$ and the filtering input $P$. Rather than processing them by two separate subnetworks, or simply concatenating them as one input to a single network, we introduce a cooperative feature extraction module based on the NAF-Block \cite{chen2022nafnet}, 
to jointly extract features from $I$ and $P$.
This module is expected to effectively capture correlations between the two sources. The detailed structure of CMFE is shown in Fig.~\ref{fig:smfe}. As shown in the figure, we first initialize features by\vspace{-2.1mm} 
\begin{gather}
    i_0 = \mathrm{GELU}(\mathrm{Conv}(I)), \\
    p_0 = \mathrm{GELU}(\mathrm{Conv}(P)),\\
    ip_0 = \mathrm{GELU}(\mathrm{Conv}(\mathrm{Cat}[I,P])),
\end{gather}
\vskip -2.1mm
\noindent where $\mathrm{Cat}[\cdot,\cdot]$ denotes the catenating operation, $\mathrm{Conv}(\cdot)$ refers to convolutional layers, and $\mathrm{GELU}(\cdot)$ refers to the Gaussian error linear unit (GELU) activation function \cite{hendrycks2016gaussian}. 
Then we build hierarchical architectures for multi-scale feature extraction as follows:\vspace{-2mm} 
\begin{gather}
    i_{t+1} = \left(\mathrm{NAF}(\mathrm{Cat}[i_t, ip_t])+i_t\right)\downarrow,\\
    p_{t+1} = \left(\mathrm{NAF}(\mathrm{Cat}[p_t, ip_t])+p_t\right)\downarrow,\\
    ip_{t+1} = \left(\mathrm{GELU}(\mathrm{Conv}(\mathrm{Cat}[i_t, p_t]))\right)\downarrow,
\end{gather}
\vskip -2mm
\noindent where $t$ refers to the scale number ranging from integer 0 to 3, $\mathrm{NAF}(\cdot)$ denotes the NAF-block, and $(\cdot)\downarrow$ denotes convolutional layers with an average pooling operator for downsampling.
The multi-scale structure here can not only save computational cost but also facilitate better feature representation for both convolution and CA-based architectures, as claimed in \cite{chen2021crossvit}. 
With the above design, the feature extractor is expected to extract shared and private informative features effectively from input image pairs.

\subsection{FeGF for feature-level guided fusion \label{sec:fegf}}
\begin{figure}[t]
	\setlength{\abovecaptionskip}{0.1cm}
	\centering
	\includegraphics[width=0.95\linewidth]{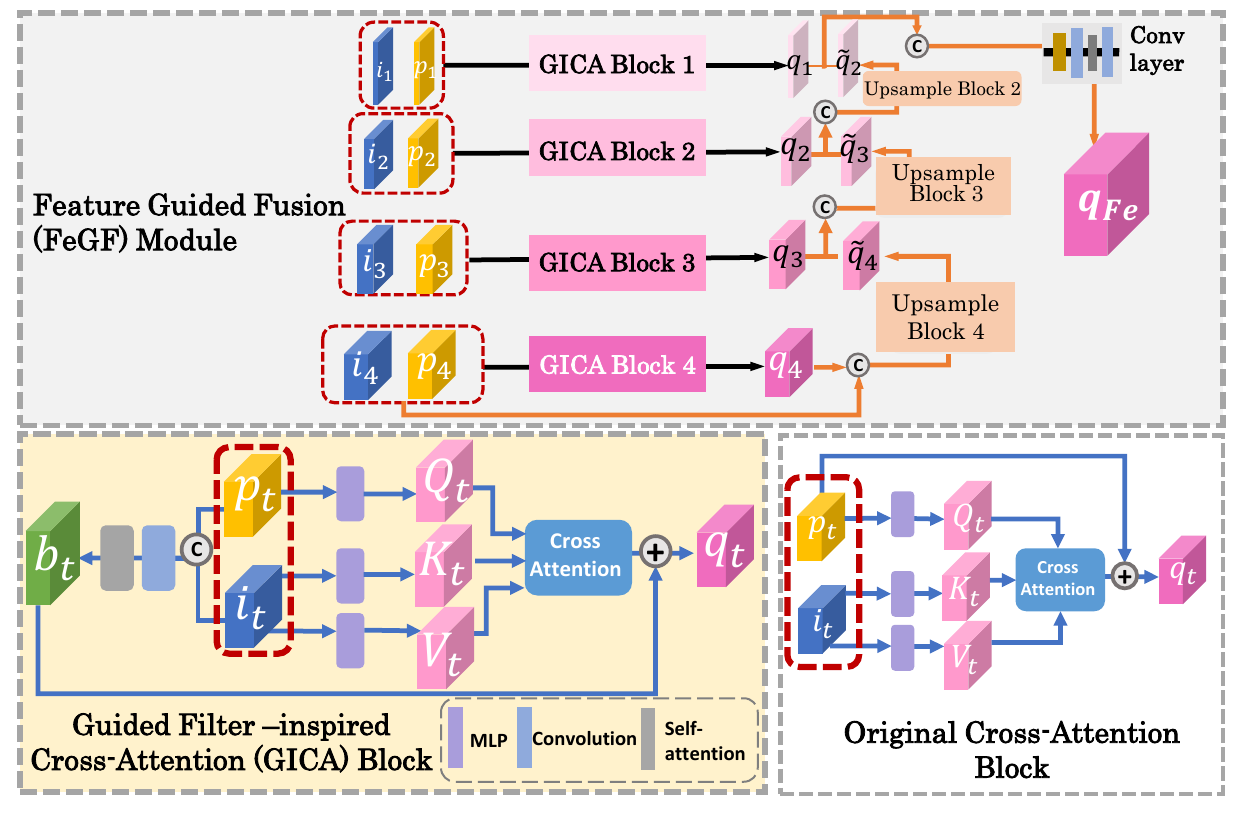}
	\vspace{-0.1cm}
	\caption{Overview of FeGF module introduced in Section~\ref{sec:fegf}.}
	\label{fig:fegf}
	\vspace{-0.6cm}
\end{figure}

The FeGF module is built upon the proposed GICA block, which inherits the principle of both CA and GF, aiming at better feature-level fusion. 
Specifically, 
as shown in Fig.~\ref{fig:fegf}, the FeGF module consists of 4 GICA blocks for fusing multi-scale features extracted by the CMFE. In each scale, the GICA block fuses features in the following way:\vspace{-2.5mm} 
\begin{gather}
    q_t = \mathrm{CA}(p_t, i_t)+b_t,\label{eq:fegf_ca}\\
    b_t = \mathrm{SA}(\mathrm{LN}(\mathrm{Cat}[p_t, i_t])) + p_t, 
\end{gather}
\vskip -2.5mm
\noindent where $\mathrm{LN}(\cdot)$ refers to layer normalization, and $\mathrm{SA}(\cdot)$ is the self-attention block.
In the above computation, we replace the original skip-connect term $p_t$ with a ``residual'' term $b_t$ to compute the fusion result $q_t$ in addition to CA (see the bottom left and bottom right plots Fig.~\ref{fig:fegf} for a comparison). According to the discussions in Section~\ref{sec:discussion}, the calculation of $q_t$ can be seen as a generalization of GF, which inherits the intrinsic mechanism of GF, while is expected to be more flexible for feature fusion. 
Besides, the calculation of $b_t$ in Eq.~\eqref{eq:fegf_ca} also follows the principle of computing $B_{GF}$, thinking $\mathrm{SA}(\mathrm{LN}(\mathrm{Cat}(p_t, i_t)))$ as $a_k\mu_k$ in Eq.~\eqref{eq:b_gf}.
To save the computational cost of CA and SA blocks, we also borrow the idea of neighborhood attention from NAT \cite{hassani2023neighborhood, hassani2022dilated} for constructing the GICA block. 

After the scale-wise fused features, $\{q_t\}_{t=1}^4$, are calculated, we aggregate them with upsampling and attention-based concatenation as follows:
\vspace{-1.8mm}
\begin{equation}
\tilde{q}_t=g_t(\mathrm{CPA}[q_t,\tilde{q}_{t+1}];\theta_g^t),~~t=1,2,3,4.
	\vspace{-1.8mm}
\end{equation}
Here, $\mathrm{CPA}[\cdot, \cdot]$ refers to the concatenation operation with a channel and space attention layer, $g_1(\cdot;\theta_g^1)$ is a small convolutional network, and $g_t(\cdot;\theta_g^t),t=2,3,4$ denotes the upsample block illustrated in Fig.~\ref{fig:fegf}. In addition, we define $\tilde{q}_{5}=[i_4,p_4]$ as the supplement to upsample $q_4$. Then the final output of the whole FeGF module is $q_{Fe}=\tilde{q}_1$.

With such a construction, our proposed FeGF module has the capability to not only retain the effective properties of CA but also inherit the advantages of the GF mechanism, which can be preliminarily observed in Fig.~\ref{fig:features}.  
It can be seen that the outputs of CA mainly depict edge information, and $b_0$ tends to preserve residual contents from the filtering input, which behaves similarly to GF in mechanism.


\subsection{ImGF for image-level guided fusion \label{sec:imgf}}

The ImGF module is designed to fuse $I$ and $P$ in a guided filtering way for better detail preservation, sharing a similar idea as \cite{pan2019spatially, wu2018fgf}. In specific, it produces the fusion result by\vspace{-2mm}
\begin{equation}
	Q_{Im} =  A_{Im} \circ I + B_{Im}, \label{eq:10}
	\vspace{-2mm}
\end{equation}
where $A_{Im}$ and $B_{Im}$ should be inferred using networks with the information of $I$ and $P$. This is expected to retain the textual information of $I$ as much as possible according to the physical model of GF discussed in Section~\ref{sec:gf}.

As discussed before, $A_{Im}$ should characterize the intrinsic correlations between $I$ and $P$, and thus it is natural to use the features extracted from them as inputs of the inference network. Besides, according to Eq.~(\ref{eq:a_gf}), the statistics of guidance $I$, i.e., variance, is adopted to normalize $A$ in GF, and therefore it is reasonable to treat $I$ as input for inferring $A_{Im}$. Combining these two observations, we can use a small convolutional neural network (CNN) to infer $A_{Im}$:\vspace{-2mm}
\begin{equation}
	A_{Im} = F_A(\{i_1,p_1,I\};\theta_A), \label{eq:imgf_a}
	\vspace{-2mm}
\end{equation}
where $F_A(\cdot;\theta_A)$ is the inference network parameterized by $\theta_A$, and $(i_1,p_1)$ denotes features output by the first CMFE block. Note that we only use the features with the same spatial size as $I$ and $P$, which makes the concatenation simpler.

According to Eq.~(\ref{eq:b_gf}), the calculation for $B$ in GF is based on $A$, $P$ and $I$. Therefore, we can use $I$ and $P$, as well as $A_{Im}$, as the input to the inference network of $B_{Im}$:
\vspace{-2mm}
\begin{equation}
	B_{Im} = F_B(\{A_{Im},P,I\};\theta_B), \label{eq:imgf_b}
	\vspace{-2mm}
\end{equation}
where $F_B(\cdot;\theta_B)$ is a small CNN parameterized by $\theta_B$.

\begin{figure}[t]
	\setlength{\abovecaptionskip}{0.1cm}
	\centering
	\includegraphics[width=1\linewidth]{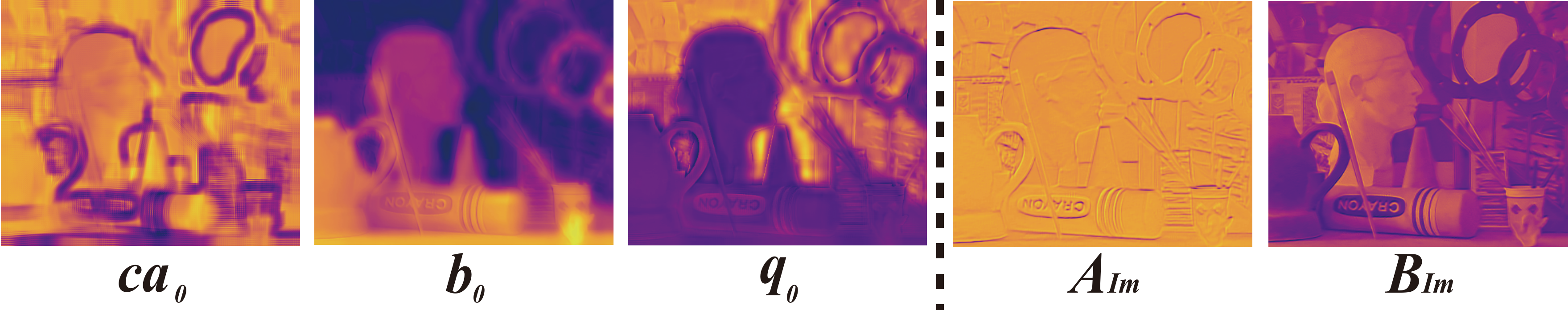}
	\vspace{-0.5cm}
	\caption{Visualization of the middle features by FeGF and ImGF.}
	\label{fig:features}
	\vspace{-0.6cm}
\end{figure}

With the designing mechanism discussed above, the learnable guided fusion in the image domain can be achieved by ImGF, as shown in Fig.~\ref{fig:features}.
It should be also noticed that though our ImGF module shares a similar idea with SVLRM \cite{pan2019spatially} by making the coefficients $A$ and $B$ in GF learnable, we have more deeply exploited the GF mechanism in the image domain. 
Specifically, SVLRM directly treats $P$ and $I$ as inputs to simultaneously predict
$A$ and $B$ without distinguishing them in learning, while we more closely follow the GF mechanism by using different source information to learn $A$ and $B$ as shown in Eqs.~(\ref{eq:imgf_a}) and (\ref{eq:imgf_b}). Consequently, our ImGF module is expected to be more powerful in emphasizing desired information by both coefficients according to the GF mechanism, and achieve better performance as demonstrated in Section~\ref{sec:ablation}.

%% file: sec/4_exp.tex
\section{Applications and Experiments}
In this section, we apply the proposed SFIGF to 4 typical GIR tasks, including GDSR, MFIF, pan-sharpening, and guided LRIE, and conduct experiments to verify its effectiveness. In addition, we also provide ablation studies to analyze the effects of each component of SFIGF. The visual results presented in this section are better viewed by zooming on a computer screen.  


\subsection{SFIGF for GDSR}
The GDSR task aims at restoring a high-resolution (HR) depth map from an LR one under the guidance of an RGB image captured in the same scene. 


\noindent\textbf{Settings.}
Following previous settings \cite{he2021towards, zhao2022dct}, we take the first 1000 paired images of the NYU V2 dataset \cite{silberman2012indoor} as the train set, and the last 449 pairs as a test set. Besides, the trained model is also tested on the Middlebury dataset \cite{hirschmuller2007mid} and the Lu \cite{lu2014depth} dataset for evaluating its generalization ability. 3 SR scales, i.e., 4x, 8x, and 16x, are considered for all datasets.
We compare our SFIGF with 6 general-purpose GIR methods, including the classical GF \cite{he2010guided, he2012guided}, and 5 deep learning-based methods, FGF \cite{wu2018fgf}, SVLRM \cite{pan2019spatially}, DKN\cite{kim2021deformable}, CU-Net\cite{deng2020cunet}, and DAGF\cite{zhong2021dagif}, and 4 state-of-the-art specialized deep learning-based GDSR approaches, including FDSR \cite{he2021towards}, DCTNet\cite{zhao2022dct},  AHMF \cite{zhong2021high}, and SSDNet \cite{zhao2023spherical}. 
The widely used root-mean-square error (RMSE) is adopted as the metric for quantitative evaluation.


\noindent\textbf{Results.}
The quantitative results of all the competing methods across all datasets are summarized in Table~\ref{tab:gdsr}. As can be seen, our SFIGF attains the leading performance across all benchmarks evaluated, indicating that it can reconstruct the depth map closest to the ground truth across different scenarios. Besides, our method also consistently achieves the lowest or competitive average RMSE at each scale, showing its robustness 
against various degrees of degradation.

\begin{figure}[t]
	\setlength{\abovecaptionskip}{0.2cm}
	\centering
	\includegraphics[width=1\linewidth]{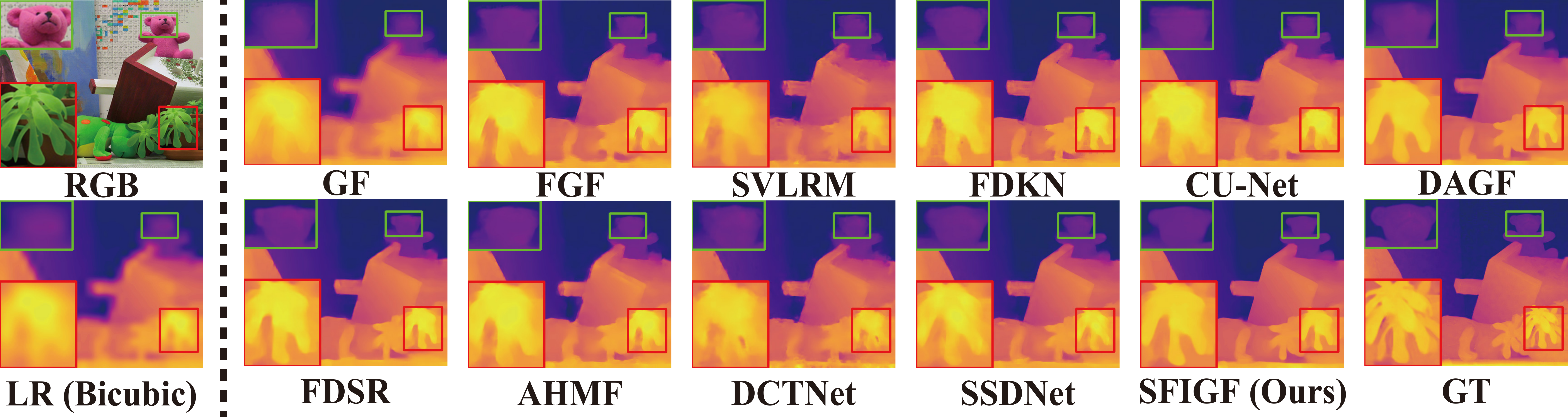}
	\vspace{-0.5cm}
	\caption{Visual results of 16x GDSR on Middlebury.}
	\label{fig:gdsr}
	\vspace{-0.3cm}
\end{figure}

The visual results of our method shown in Fig.~\ref{fig:gdsr} are also promising.
Compared with the methods that are implemented with image-level guided filtering mechanisms, including GF \cite{he2010guided}, FGF \cite{wu2018fgf}, and SVLRM \cite{pan2019spatially}, our SFIGF avoids artifacts (like ghosting) and preserves structures, simultaneously. When competing against other general deep learning-based GIR approaches like FDKN\cite{kim2021deformable}, CU-Net \cite{deng2020cunet}, and DAGF \cite{zhong2021dagif}, our method better reconstructs fine-grained texture details. Compared with the specialized GDSR method, e.g. FDSR \cite{guo2018hierarchical}, AHMF\cite{zhong2021high}, DCTNet\cite{zhao2022dct}, SSDNet \cite{zhao2023spherical}, our method produces smoother surfaces and clearer structural details. These results demonstrate that our method can not only effectively mitigate the limitations of general-purpose GIR methods, but also beat the current state-of-the-art deep models for the GDSR task.

\begin{table}[t]
	\setlength\tabcolsep{3pt}
	\centering
	\caption{Quantitative results (RMSE) of the GDSR task. ``Average'' means the RMSE averaged over all images across datasets.}
	\vspace{-2mm}
	\begin{adjustbox}{width=\columnwidth,center}
		\begin{tabular}{ccccccccccccc}
			\toprule
			\multirow{2}{*}{Methods}& \multicolumn{3}{c}{Middlebury} & \multicolumn{3}{c}{NYU V2} &  \multicolumn{3}{c}{Lu} & \multicolumn{3}{c}{Average}\\ 
			\cmidrule(lr){2-4}
			\cmidrule(lr){5-7}
			\cmidrule(lr){8-10}
			\cmidrule(lr){11-13}
			~ & 4x & 8x & 16x & 4x & 8x & 16x & 4x & 8x & 16x & 4x & 8x & 16x \\ \hline
			GF\cite{he2010guided}& 3.13 & 4.00 & 6.19 & 5.97 & 7.40 & 11.2 & 4.20 & 5.21 & 7.46 & 5.79 & 7.19 & 10.9
			\\
			FGF \cite{wu2018fgf} & 1.36 & 2.28 & 4.34 & 2.39 & 3.82 & 6.98 & 1.37 & 2.35 & 4.61 & 2.32 & 3.72 & 6.81 
			\\ 
			SVLRM \cite{pan2019spatially} & 1.18 & 2.43 & 5.33 & 1.82 & 3.67 & 7.87 & 1.04 & 2.51 & 6.15 & 1.78 & 3.59 & 7.71 
			\\ 
			FDKN \cite{kim2021deformable} & 1.09 & 2.09 & 4.27 & 1.64 & 3.30 & 6.62 & 1.02 & 2.07 & 4.83 & \underline{1.60} & 3.22 & 6.47 
			\\ 
			CU-Net \cite{deng2020cunet}& 1.28 & 2.40 & 4.73 & 2.22 & 4.02 & 7.88 & 1.05 & 2.48 & 5.19 & 2.15 & 3.91 & 7.67 
			\\ 
			DAGF \cite{zhong2021dagif} & 1.28 & 2.66 & \underline{3.80} & 2.66 & 4.97 & 7.19 & 1.26 & 2.60 & 4.77 & 2.57 & 4.81 & 6.97 
			\\ \hline
			FDSR \cite{he2021towards} & 1.06 & 2.08 & 4.17 & 1.78 & 3.17 & 5.94 & 1.15 & 2.19 & 5.08 & 1.73 & 3.10 & 5.83 
			\\ 
			AHMF \cite{zhong2021high} & 1.19 & 2.35 & 4.76 & 1.83 & 3.74 & 7.16 & 0.96 & 2.33 & 5.38 & 1.78 & 3.65 & 7.01 
			\\
			DCTNet \cite{zhao2022dct} & 1.10 & 2.05 & 4.19 & 1.73 & 3.56 & 6.50 & 0.88 & 1.85 & \underline{4.39} & 1.69 &3.46 &	6.35 \\ 
			SSDNet \cite{zhao2023spherical} & \underline{1.02} & \underline{1.91} & 4.02 & \textbf{1.60} & \underline{3.14} & \underline{5.86} & \textbf{0.80} & \underline{1.82} & 4.77 & \textbf{1.56} & \underline{3.06} & \underline{5.74} 
			\\ \hline
			SFIGF (Ours) & \textbf{1.01} & \textbf{1.74} &\textbf{3.38}& \underline{1.68} & \textbf{3.05} & \textbf{5.75} & \textbf{0.80} &\textbf{1.75} & \textbf{4.38} & 1.63 & \textbf{2.96} & \textbf{5.60} \\ 
   \toprule
		\end{tabular}
	\end{adjustbox}
	\label{tab:gdsr}
	\vspace{-6mm}
\end{table}

\subsection{SFIGF for pan-sharpening}
The pan-sharpening problem focuses on restoring a high-resolution multi-spectral (HRMS) image from a low-resolution multi-spectral (LRMS) image with the help of an HR panchromatic image.


\noindent\textbf{Settings.}
We use the WorldView-\uppercase\expandafter{\romannumeral3} dataset \cite{deng2022machine} for training and testing.
Similar to the GDSR experiments, we consider both general-purpose GIR methods and specialized methods. Specifically, we compare our method with the same set of general-purpose methods mentioned before, and 4 representative methods for the pan-sharpening task, including LAGConv \cite{jin2022lagconv}, PANNet \cite{yang2017pannet}, GPPNN \cite{xu2021gppnn}, and MADUNet \cite{madunet}. 
We adopt 6 metrics in this task: PSNR and SSIM \cite{wang2004image} are general image quality assessment (IQA) metrics; 
the spectral angle mapper (SAM) \cite{yuhas1992discrimination} assesses the spectral similarity for hyperspectral images (HSIs);
the relative dimensionless global error in synthesis (ERGAS) and correlation coefficient (SCC) capture error magnitudes across spectral bands; and the Q-index \cite{vivone2014critical} measures image quality comprehensively. 

\begin{table}[t]
	\setlength\tabcolsep{3pt}
	\centering
	\caption{Quantitative results of pan-sharpening on WorldView-III.}
	\vspace{-2.5mm}
	\begin{adjustbox}{width=0.88\columnwidth,center}
		\begin{tabular}{ccccccc}
			\toprule
			Methods & PSNR$\uparrow$  &SSIM$\uparrow$ & SAM$\downarrow$ & ERGAS$\downarrow$ & SCC$\uparrow$ & Q$\uparrow$\\ 
			\hline
			GF \cite{he2012guided} & 27.78 & 0.805 & 0.125 & 6.542 & 0.893 & 0.497  \\
			FGF \cite{wu2018fgf} & 32.49	&0.920	&0.086	&3.971	&0.947	&0.722\\
			SVLRM \cite{pan2019spatially}&32.99&0.944	&0.092	&3.572	&0.959	&0.755 \\
			DKN \cite{kim2021deformable} & 27.46 &0.742 & 0.107 & 6.861 & 0.841 & 0.418 \\ 
			CU-Net \cite{deng2020cunet} & 32.46	&0.927	&0.101	&3.993	&0.949	&0.736\\ 
			DAGF \cite{zhong2021dagif} & 31.57 & 0.886&0.117&4.444&0.937&0.684\\
			\hline
			LAGConv \cite{jin2022lagconv} & \underline{37.12} & \underline{0.972}	& \underline{0.053}	& \underline{2.264}	& \underline{0.980} &	\underline{0.823}\\
			PANNet\cite{yang2017pannet} &32.47	&0.913	&0.097	&3.927	&0.951	&0.724\\
			GPPNN \cite{xu2021gppnn} & 35.55 &	0.962	&0.065	& 2.719 &	0.973	& 0.794\\
			MADUNet \cite{madunet} & \underline{37.12}	&0.971	&0.054	&2.267	&\textbf{0.981}	& \underline{0.823} \\
			\hline
			SFIGF (Ours) & \textbf{37.30} &	\textbf{0.973}	&\textbf{0.052}	&\textbf{2.215}	&\textbf{0.981}	& \textbf{0.830} \\ \hline
		\end{tabular}
	\end{adjustbox}
	\label{tab:pan}
	\vspace{-6mm}
\end{table}

\begin{figure*}[t]
	\setlength{\abovecaptionskip}{0.2cm}
	\centering
	\includegraphics[width=1\linewidth]{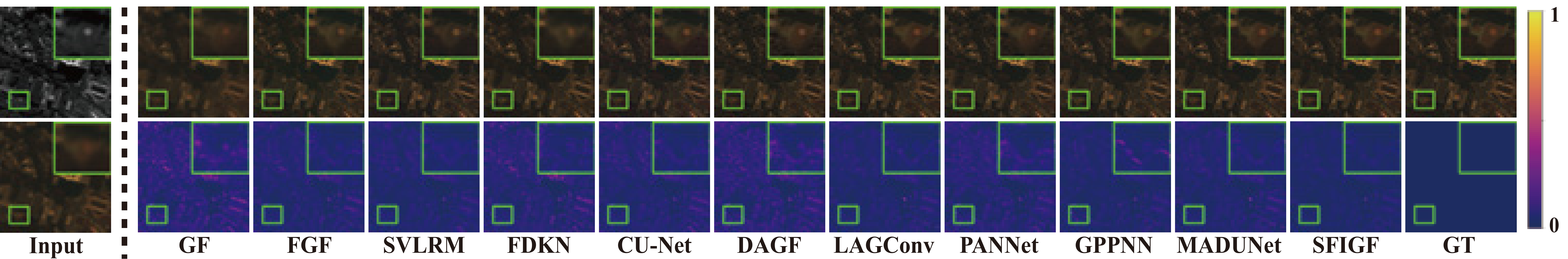}
	\vspace{-0.50cm}
	\caption{Visual results of pan-sharpening on  WorldView-\uppercase\expandafter{\romannumeral3}. Top: the restored images. Bottom: the corresponding RMSE maps.}
	\label{fig:pan}
	\vspace{-0.5cm}
\end{figure*}

\noindent\textbf{Results.}
The quantitative results are shown in Table~\ref{tab:pan}. As can be seen, our SFIGF ranks first with respect to all the metrics, significantly outperforming the general-purpose GIR methods and also performing competitively against the state-of-the-art pan-sharpening methods. Specifically, the highest PSNR and SSIM values demonstrate the strong ability of SFIGF in preserving spatial information; the lowest SAM and ERGAS values indicate its ability in accurately retaining the spectral features and spatial patterns of the source data; the highest SCC rating shows its effectiveness in maintaining local cross-band correlations; and the leading score in Q-index suggests its capacity in synthesizing results most consistent with human perception.
These results quantitatively verify the effectiveness of the proposed SFIGF for the pan-sharpening task.

The visual results in Fig.~\ref{fig:pan} further substantiate the effectiveness of SFIGF.
As shown in the top row, our SFIGF is able to successfully reconstruct fine-scale details that degraded in the LRMS images. The RMSE maps in the bottom row provide a more easily observable validation of the qualitative findings, that the prediction by SFIGF most closely matches the ground truth HRMS images.
These results suggest our method can effectively leverage both textural and contextual information to reconstruct high-frequency details for this task.

\subsection{SFIGF for MFIF}
The MFIF task aims at fusing spatially complementary visual contents from multiple images exhibiting focus disparities and reconstructing an all-in-focus image. Typically in this task, two inputs depicting the same scene at near and far focal distances are fused. 
Therefore, this task can be treated as a mutual GIR, where the two inputs guide each other for the final result. Besides, this task is commonly formulated in an unsupervised manner to meet practical requirements, and thus we follow this setting. The detailed network and unsupervised training losses for this task are provided in the supplementary material (SM) due to page limitation.

\begin{table}[t] \footnotesize
	\centering
	\setlength{\abovecaptionskip}{0.1cm}
	\caption{Quantitative results of MFIF on REALMFF and Lytro.}
	\begin{adjustbox}{width=0.95\columnwidth,center}
		\begin{tabular}{cccccc}
			\toprule
			\multirow{2}{*}{Methods}& \multicolumn{3}{c}{RealMFF} & \multicolumn{2}{c}{Lytro} \\ 
			\cmidrule(lr){2-4}
			\cmidrule(lr){5-6}
			~ & PSNR$\uparrow$  &SSIM$\uparrow$ & LPIPS$\downarrow$ & Q$_{MI}$$\uparrow$ & Q$_{S}$$\uparrow$ \\
			\hline
			GF \cite{he2012guided} &31.88&  0.918 & 0.098 & 1.8262 & 0.8408 \\
			FGF \cite{wu2018fgf} & 37.34 & 0.967 & \textbf{0.008} & 1.8675 & 0.8507 \\
			SVLRM \cite{pan2019spatially}&36.79& 0.975 & 0.014 & 1.8873 & \underline{0.8521} \\
			DKN \cite{kim2021deformable} & 34.22& 0.942 & 0.080 & 1.3742 & 0.8386\\ 
			CU-Net \cite{deng2020cunet} & \underline{38.68} & 0.975 & 0.011 &1.8614 & 0.8410 \\ 
			DAGF \cite{zhong2021dagif} & 37.69 & 0.975 & 0.015 & 1.8961 & 0.8414 \\
			\hline
			IFCNN \cite{liu2017multi} & 31.61 &0.914 & 0.034 & 1.5769 & 0.7874\\
			SMFUSE \cite{ma2021smfuse} &38.54 & \underline{0.977} & 0.019 & 1.8687 & 0.8404 \\
			U2Fusion \cite{xu2020fusiondn} & 35.76 &0.972 & 0.047 & 1.8734 & 0.8318 \\
			ZMFF \cite{hu2023zmff} &34.88 & 0.956 & 0.028 & \underline{1.8874} & 0.8365  \\
			\hline
			SFIGF (Ours) & \textbf{39.71}&	\textbf{0.986} & \underline{0.009} & \textbf{1.9160} & \textbf{0.8545} \\
   \toprule
		\end{tabular}
	\end{adjustbox}
	\label{tab:mfif}
	\vspace{-6mm}
\end{table}

\noindent\textbf{Settings.}
We conduct experiments 
on the Real-MFF dataset \cite{zhang2020real}, which consists of various natural multi-focus images with ground truth, generated by light field images.
We use the first 650 pairs of images for training and the last 60 pairs for testing. 
Besides, we also evaluate the trained model 
on the 
real-world Lytro dataset \cite{nejati2015multi} to show its generalization ability. 
For comparison, in addition to the 6 general-purpose GIR methods used before, 4 representative deep learning-based MFIF methods are adopted, including IFCNN \cite{liu2017multi}, SMFUSE \cite{ma2021smfuse}, U2Fusion \cite{xu2020fusiondn}, and ZMFF \cite{hu2023zmff}. All the competing methods are trained in an unsupervised manner or implemented in a zero-shot way, without reference to the ground truth.
We adopt 3 IQA metrics, including PSNR, SSIM, and LPIPS \cite{zhang2018lpips}, for evaluation on the Real-MFF dataset. LPIPS is a deep feature-based metric for assessing the perceptual quality of an image with a reference one.
For the real-world Lytro dataset without ground truth, we follow \cite{hu2023zmff} and adopt metrics Q$_{MI}$ and Q$_{S}$, which respectively measure the average mutual information and SSIM value between the fusion result and two sources.

\begin{figure}[t]
	\setlength{\abovecaptionskip}{0.1cm}
	\centering
	\includegraphics[width=1\linewidth]{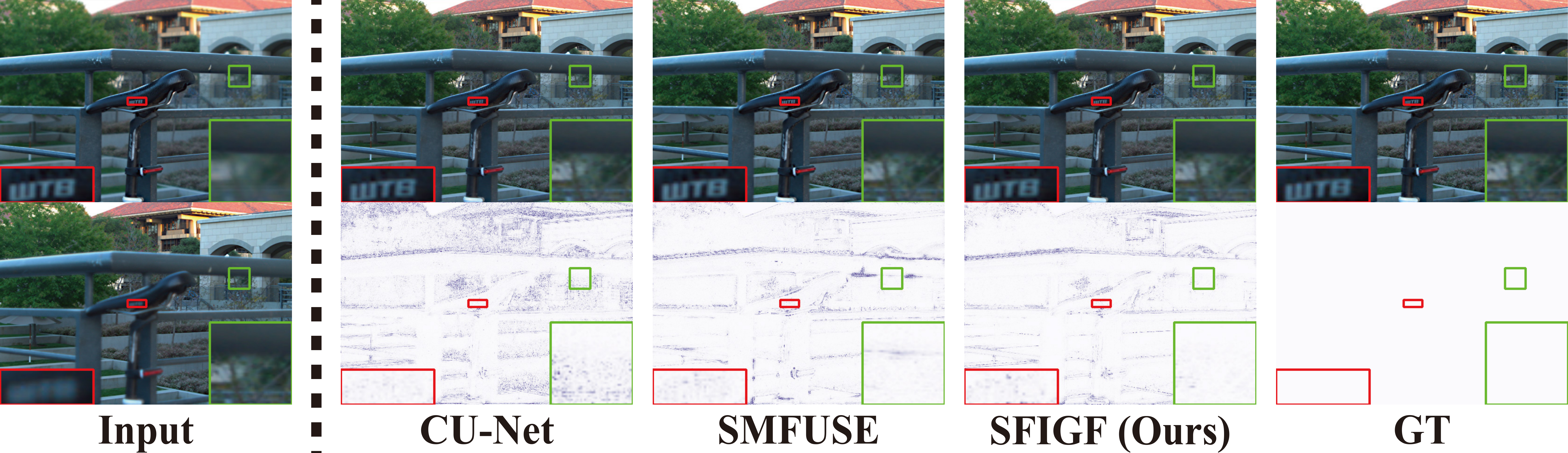}
	\caption{Visual results of MFIE on Real-MEF. Top: the restored images. Bottom: the corresponding error maps.}
	\label{fig:mff}
	\vspace{-0.5cm}
\end{figure}

\noindent\textbf{Results.}
The quantitative results of all competing methods are reported in Table~\ref{tab:mfif}, and we can see that our SFIGF achieves the best performance among all methods, obtaining 1.03 dB improvement in PSNR and 0.009 improvement in SSIM compared with the second-best ones, and takes the second place with respect to LPIPS. 
On the Lytro dataset, we also achieve the best quantitative performance, 
indicating that our results retain the richest information from source images, and are also most similar to sources in structural details. Fig.~\ref{fig:mff} shows example visual results on the Real-MFE dataset. Due to space limitations, we only show results of the most competitive methods and more visualizations are in the SM. The bottom images are the error maps with respect to the ground truth, following \cite{hu2023zmff}, for a clearer comparison. It can be seen that our method produces results with fewer residuals on both far and near focus areas, which further demonstrates its effectiveness.

\subsection{SFIGF for guided LRIE}

Following the work by Xiong et al. \cite{xiong2021mffnet} and Dong et al. \cite{dong2022abandoning}, we consider two kinds of guidance, i.e., the red flashlight image \cite{xiong2021mffnet} and the monochrome image \cite{dong2022abandoning}, respectively, for enhancing the low-light RAW image.


\noindent\textbf{Settings.}
For the guided LIRE with red flashlight, we synthesize paired RAW/red flashlight data from the Sony subset of the SID dataset \cite{chen2018learning}. 
For the guided LIRE with monochrome image, we adopt the Mono-Colored RAW (MCR) paired dataset collected by Dong et al. \cite{dong2022abandoning}. 
In addition to the 6 general-purpose GIR methods, we consider 2 recently proposed guided LRIE methods, i.e., MFFNet \cite{xiong2021mffnet} and DBLE \cite{dong2022abandoning}.
4 metrics, including PSNR, SSIM, LPIPS and Delta E ($\triangle E^{*}$) \cite{backhaus2011color}, are adopted for quantitative evaluation, where
$\triangle E^{*}$ is a metric to measure color distortion.

\begin{figure}[t]
	\setlength{\abovecaptionskip}{0.1cm}
	\centering
	\includegraphics[width=1\linewidth]{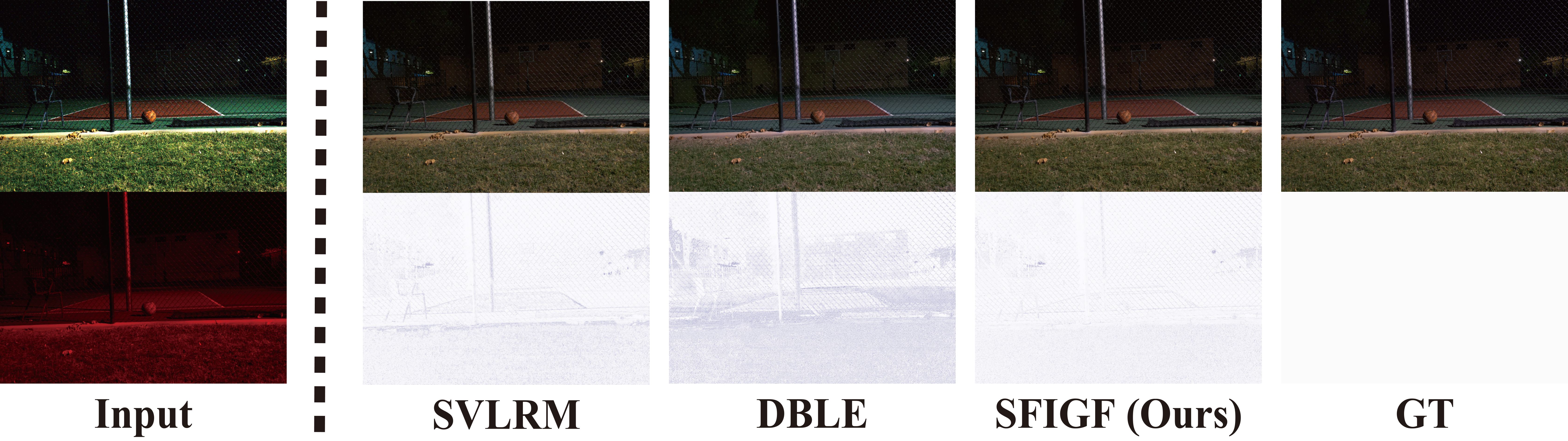}
	\vspace{-0.4cm}
	\caption{Visual results of guided LIE on SID-Sony. Top: the restored images. Bottom: the corresponding error maps.}
\label{fig:lrie}
\vspace{-0.35cm}
\end{figure}

\noindent\textbf{Results.}
The quantitative results for all competing methods are summarized in Table~\ref{tab:glrie}. It can be seen that our proposed SFIGF model achieves superior performance with respect to all adopted metrics, 
which shows SFIGF has outstanding abilities in reconstructing textual and contextual information and also alleviating color biases. Fig.~\ref{fig:lrie} shows example visual results of the most competitive methods on the SID-Sony dataset, and more results are in the SM. 
It presents that our method produces the result with more details while fewer color biases, further confirming its superiority.

 \subsection{Effectiveness of modules in SFIGF}\label{sec:ablation}
In this subsection, we conduct experiments on the 16x GDSR task to verify the necessity and effectiveness of key modules in SFIGF, including CMFE, ImGF, and FeGF. The overall results are summarized in Table~\ref{tab:ablation} and Fig.~\ref{fig:ablation}. 


\noindent\textbf{Effectiveness of the CMFE module.}
We first conduct experiments to evaluate the effectiveness of the CMFE module in cooperatively extracting shared and private information from source images. Specifically, we compared the full SFIGF against two baselines\textemdash a \emph{shared} setting where source images are directly concatenated as input, denoted as CMFE$_{s}$, and a \emph{private} setting where sources are processed independently by separate feature extractors, denoted as CMFE$_{p}$.
For CMFE$_{s}$, the concatenated inputs are passed through a feature extractor with NAF blocks, keeping the skip connections. In CMFE$_{p}$, two sources are processed by isolated NAF feature extractors.

\begin{table}[t]
\setlength\tabcolsep{3pt}
\centering
\setlength{\abovecaptionskip}{0.1cm}
\caption{Quantitative results of the guided LIRE task.}
\vspace{-1mm}
\begin{adjustbox}{width=\columnwidth,center}
	\begin{tabular}{ccccccccc}
		\toprule
		\multirow{2}{*}{Methods}&  \multicolumn{4}{c}{SID-Sony (RAW+Red)} &  \multicolumn{4}{c}{MCR (RAW+Mono)} \\ 
		\cmidrule(lr){2-5}
		\cmidrule(lr){6-9}
		~ & PSNR $\uparrow$ & SSIM $\uparrow$ & LPIPS $\downarrow$ & $\triangle E^{*}\downarrow$ & PSNR$\uparrow$ & SSIM$\uparrow$ & LPIPS$\downarrow$ & $\triangle E^{*}\downarrow$   \\ \hline
		GF \cite{he2012guided} & 16.25 & 0.674 & 0.328 & 18.61 & 16.04 & 0.697 & 0.405 & 14.47  \\
		FGF \cite{wu2018fgf}  & 30.80  & 0.857  & 0.129  & 4.453 & 23.93  & 0.830  & 0.176 & 6.660  \\
		SVLRM \cite{pan2019spatially}&32.34 & 0.890 & 0.075 & 3.857 & 25.83	& 0.879& 0.103 & 5.758 \\
		DKN \cite{kim2021deformable} & 26.68  & 0.720  & 0.232 & 6.467 & 23.43  & 0.838  & 0.151  & 7.528 \\ 
		CU-Net \cite{deng2020cunet} & 30.99  & 0.881  & 0.091 & 4.290 & 19.70  & 0.673  &  0.257 & 11.55 \\ 
		DAGF \cite{zhong2021dagif} & 27.96 &0.887 &0.090 & 5.592 & 26.67 &0.873&0.126&5.241\\
		\hline
		DBLE \cite{dong2022abandoning}  & \underline{33.36}  & \underline{0.893} & \underline{0.075}  & \underline{3.622} & \underline{31.69}  & \underline{0.908}  & \underline{0.082} &\underline{3.133}  \\ 
		MFFNet \cite{xiong2021mffnet}  & 30.99  & 0.876  & 0.101 & 4.440 & 28.62  & 0.902  & 0.096 & 4.316  \\
		\hline
		SFIGF (Ours) &\textbf{34.25}  & \textbf{0.902}  & \textbf{0.067} & \textbf{3.304} & \textbf{32.11} 	&\textbf{0.925} &	\textbf{0.065} &	\textbf{3.077}  \\ \hline
	\end{tabular}
\end{adjustbox}
\label{tab:glrie}
\vspace{-6.5mm}
\end{table}

As shown in Table~\ref{tab:ablation}, in both settings, the performance decreases compared with the full model.
Visual results in Fig.~\ref{fig:ablation} reveal that the CMFE module is able to better capture fine-grained details like object edges and produce images with a more even appearance, as compared with CMFE$_{s}$ and CMFE$_{p}$. 
Besides, we visualize the extracted features by different approaches in the SM, which further demonstrates the effectiveness of the CMFE module. 

\noindent\textbf{Effectiveness of the FeGF module.}
To demonstrate its effectiveness and analyze its mechanism, we conduct experiments with 3 variants of the FeGF and keep the model sizes similar: (c) replacing the whole FeGF with convolutional blocks, (d) realizing GF mechanism by $q_t = a_t \circ i_t + b_t$ with $a_t$ and $b_t$ being learned by convolutional layers, and (e) replacing the GICA with the original CA. 

Several findings can be drawn for the results in Table~\ref{tab:ablation} and Fig.~\ref{fig:ablation}. First, the performance of (c) decreases, which can be attributed to the lack of the structure-preserving ability of the GF mechanism.
Second, (d) outperforms (c) with sharper edges since GF mechanism has been introduced; but is worse than the full SFIGF with fewer clear details 
due to the absence of CA, which can capture long-range dependencies for better reconstructing information degraded by downsampling operations. 
Third, (e) outperforms (d) and (c) but is worse than the full SFIGF due to its incomplete GF mechanism. 
All these findings substantiate the reasonability of the CA-based GF mechanism realized by the proposed GICA block and FeGF module.

\begin{figure}[t]
	\setlength{\abovecaptionskip}{0.1cm}
	\centering
	\includegraphics[width=1\linewidth]{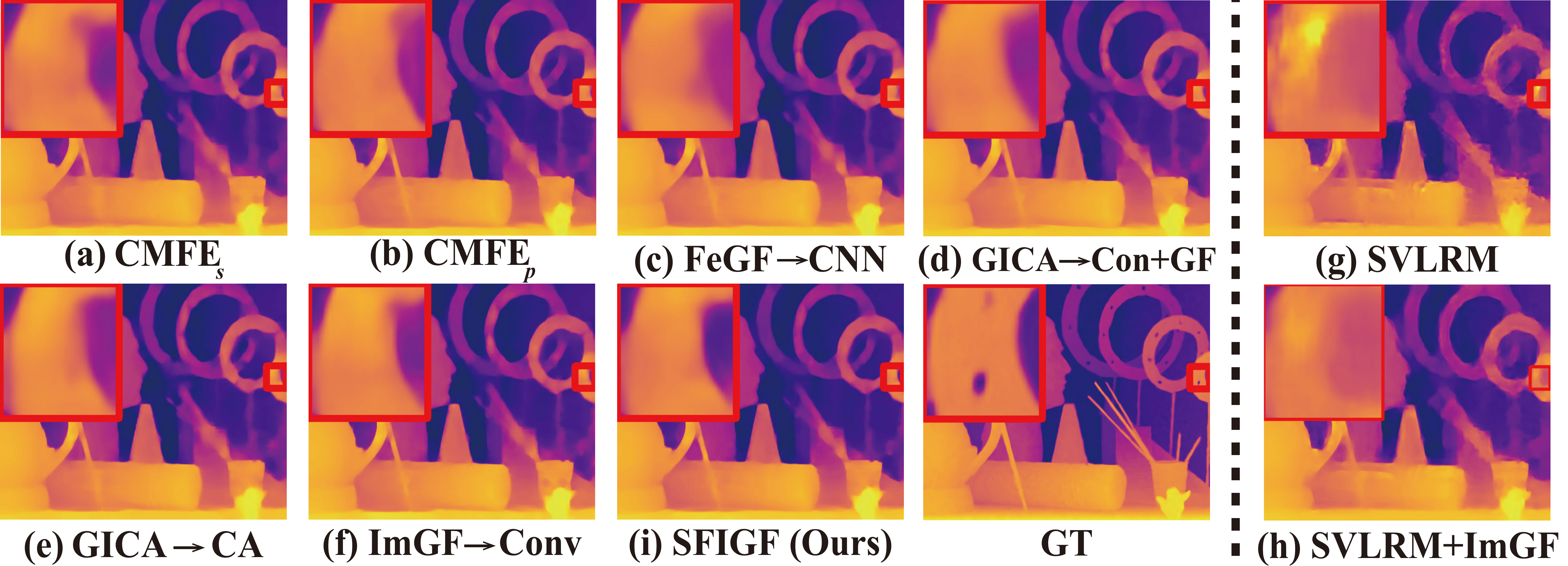}
	\caption{Visual results of the ablation study with 16x GDSR.}
	\label{fig:ablation}
	\vspace{-0.3cm}
\end{figure}

\begin{table}[t]
	\setlength\tabcolsep{3pt}
	\centering
	\caption{RMSE results of ablation study with 16x GDSR task.} \label{tab:ablation} 
	\vspace{-3mm}
	\begin{adjustbox}{width=0.95\columnwidth,center}
		\begin{tabular}{l cccc}
			\toprule
			Settings & Middlebury &NYU V2 &Lu &Average \\ \hline
			(a) CMFE$_p$   &  3.53 & \underline{5.89} &4.57& 	\underline{5.73}\\ 
			(b) CMFE$_s$  & 3.61 & 6.10 &4.45 & 5.95\\
			\hline
			(c) FeGF $\rightarrow$ CNN &3.63 	& 6.12& 4.62& 5.96\\
			(d) GICA $\rightarrow$ Conv+GF & 3.48 &	6.01& \textbf{4.29} & 5.85\\
			(e) GICA $\rightarrow$ CA & \underline{3.44} &	5.95 & 4.48 & 5.80  \\
			\hline
                (f) ImGF $\rightarrow$ Conv & 3.52 	& \underline{5.89} &4.53 & 5.79 \\
			(g) SVLRM & 5.33 & 7.87 & 6.15 & 7.71\\
			(h) SVLRM with ImGF & 4.79 & 7.28 & 5.57 & 7.12\\
			\hline
			(i) SFIGF (Ours) & \textbf{3.38} &	\textbf{5.75} & \underline{4.38} & \textbf{5.60 }\\\hline
		\end{tabular}
	\end{adjustbox}
	\vspace{-5.5mm}
\end{table}

\noindent\textbf{Effectiveness of the ImGF module.} 
We first experiment by replacing the ImGF module with CNN structures without the GF mechanism. In this setting, since no GF mechanism is implemented at the image level, the whole network indeed only realizes feature-level fusion.  As shown in (f) of Table~\ref{tab:ablation}, this leads to degraded quantitative performance. Besides, the visual result in Fig.~\ref{fig:ablation} (f) is also with fewer fine image details. These results demonstrate the necessity of image-level fusion by the ImGF module.



As discussed in Section~\ref{sec:imgf}, our ImGF shares a similar idea with SVLRM \cite{pan2019spatially}, but more coincides with the original GF in mechanism. Therefore, we conduct an experiment by introducing our learning strategy for $A_{Im}$ and $B_{Im}$ to SVLRM. The comparison results are shown in (g) and (h) of Table~\ref{tab:ablation} and Fig. \ref{fig:ablation}, and the better performance of our strategy can be clearly observed, which further substantiates the effectiveness of the designing mechanism of the ImGF.



%% file: sec/5_conclusion.tex
\section{Conclusion}
In this work, we have proposed the SFIGF for the GIR tasks, by simultaneously implementing guided fusion in feature and image domains. In the feature domain, SFIGF can effectively implement feature-level fusion by the GF-inspired GICA module; and in the image domain, SFIGF realizes image-level fusion closely following the GF mechanism. Consequently, the guided restoration result better retains both contextual and textural information extracted from source images. Experiments on 4 typical GIR tasks, including GDSR, pan-sharpening, MFIF and guided LRIE, have verified the effectiveness of the proposed method, and demonstrated its general availability.


%% file: sec/X_suppl.tex
\clearpage
\setcounter{page}{1}
\setcounter{figure}{0}
\setcounter{table}{0}
\setcounter{section}{0}
\twocolumn[{%
	\renewcommand\twocolumn[1][]{#1}%
	\maketitlesupplementary
	\begin{center}
		\centering
		\captionsetup{type=figure}
		\includegraphics[width=\linewidth]{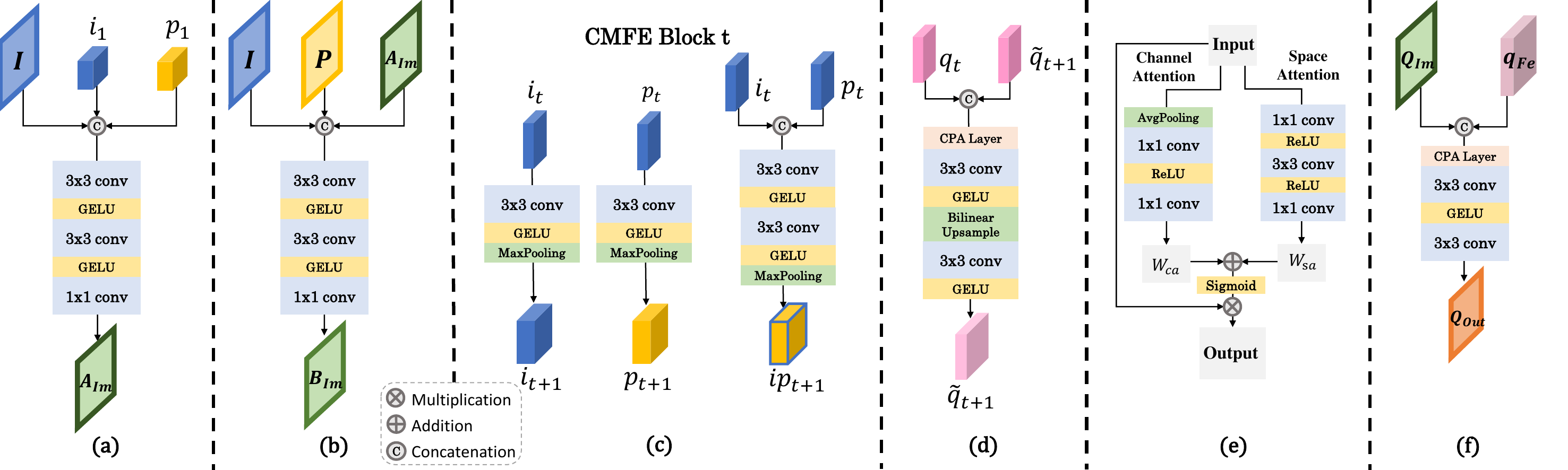}
		\captionof{figure}{Details of network blocks. (a) is the convolutional block to infer the map $A_{Im}$ in the ImGF module; (b) is the convolutional block to infer the map $B_{Im}$ in the ImGF module;  (c) is the downsampling and channel expansion operations in the CMFE module; (d) is the aggregating and upsampling operations in the FeGF module; (e) is the detailed architecture of the channel and space attention (CPA) block; (f) is the convolutional layer for aggregating  $Q_{Im}$ and $q_{Fe}$. \label{fig:supnet}}
	\end{center}%
}]


\begin{table}[t]
	\centering
	\tabcolsep=1mm
	\caption{Number of channels of convolutional layers and blocks in our SFIGF.  $n$ refers to the number of base channels. The ``Output'' denotes the convolutional layer to aggregate Q$_{Im}$ and q$_{Fe}$, as is shown in Fig.~\ref{fig:supnet} (f). \label{tab:ch}}
	\small
	\begin{tabular}{c ccc}
		\toprule
		Module    &   Block           & Num of Ch$_{in}$ & Num of Ch$_{out}$ \\
		\hline
		\multirow{8}{*}{CMFE}& Initial conv layer      & C$_{in}$& $n$\\
		& CMFE Block 1 & $n$ & $n$\\
		& Downsample 1 & $n$ & $2n$\\
		& CMFE Block 2 & $2n$ & $2n$\\
		& Downsample 2 & $2n$ & $4n$\\
		& CMFE Block 3  & $4n$ & $4n$\\
		& Downsample 3 & $4n$ & $8n$\\
		& CMFE Block 4 & $8n$ & $8n$ \\\hline
		\multirow{2}{*}{ImGF}   & F$_{A}$         & C$_{in}$$+n+n$& C$_{in}$\\
		& F$_{B}$         & C$_{in}$$+$C$_{in}$$+n$& C$_{in}$\\\hline
		\multirow{7}{*}{FeGF}   & FeGF Block 1 & $2n$& $n$\\
		& FeGF Block 2 & $4n$& $2n$\\
		& FeGF Block 3 & $8n$& $4n$\\
		& FeGF Block 4 & $16n$& $8n$\\
		& Upsample Block 2 & $2n\times2$& $2n$\\
		& Upsample Block 3 & $4n\times2$& $4n$\\
		& Upsample Block 4 & $8n\times2$& $8n$\\\hline
		\multirow{2}{*}{Output} & conv1        & C$_{in}$$+n$& C$_{in}$$+n$\\
		& conv2        & C$_{in}$$+n$& C$_{out}$\\
  \toprule
	\end{tabular}
	\normalsize
\end{table}

\section{More details of SFIGF}

In this section, we present more details of our SFIGF network. Fig.~\ref{fig:supnet} illustrates the detailed structures of network modules that are not shown in the main text, including the networks for inferring $A_{Im}$ and $B_{Im}$ in the ImGF module, the downsampling operation in the CMFE module, the aggregating and upsampling operation in the FeGF module, the channel and space attention (CPA) block, and the convolutional layers for aggregating the fusion results $Q_{Im}$ and $q_{Fe}$. We also summarize in Table~\ref{tab:ch} the number of channels of convolutional layers within our SFIGF. Note that the number of input and output channels depends on the format of images, which varies with different tasks, as shown in Section \ref{sec:sup_exp} of this supplementary meterial.

\section{More evaluations for the effectiveness of modules in SFIGF}


\subsection{Effectiveness of the CMFE module}
\begin{figure}[t]
	\setlength{\abovecaptionskip}{0.2cm}
	\centering
	\includegraphics[width=1\linewidth]{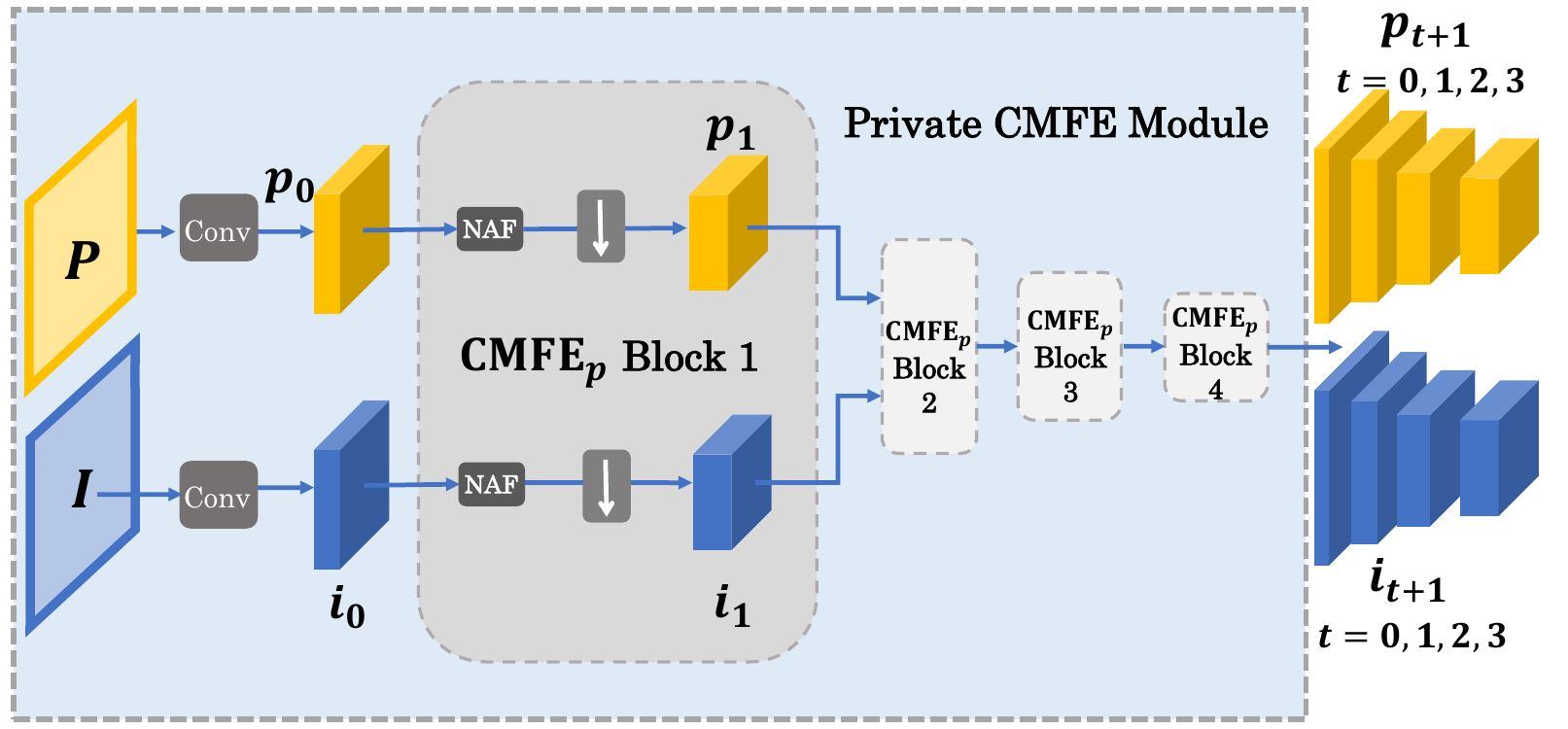}
	\caption{The detailed structure of the CMFE$_p$.}
	\label{fig:cmfe_p}
\end{figure}

\begin{figure}[t]
	\setlength{\abovecaptionskip}{0.2cm}
	\centering
	\includegraphics[width=1\linewidth]{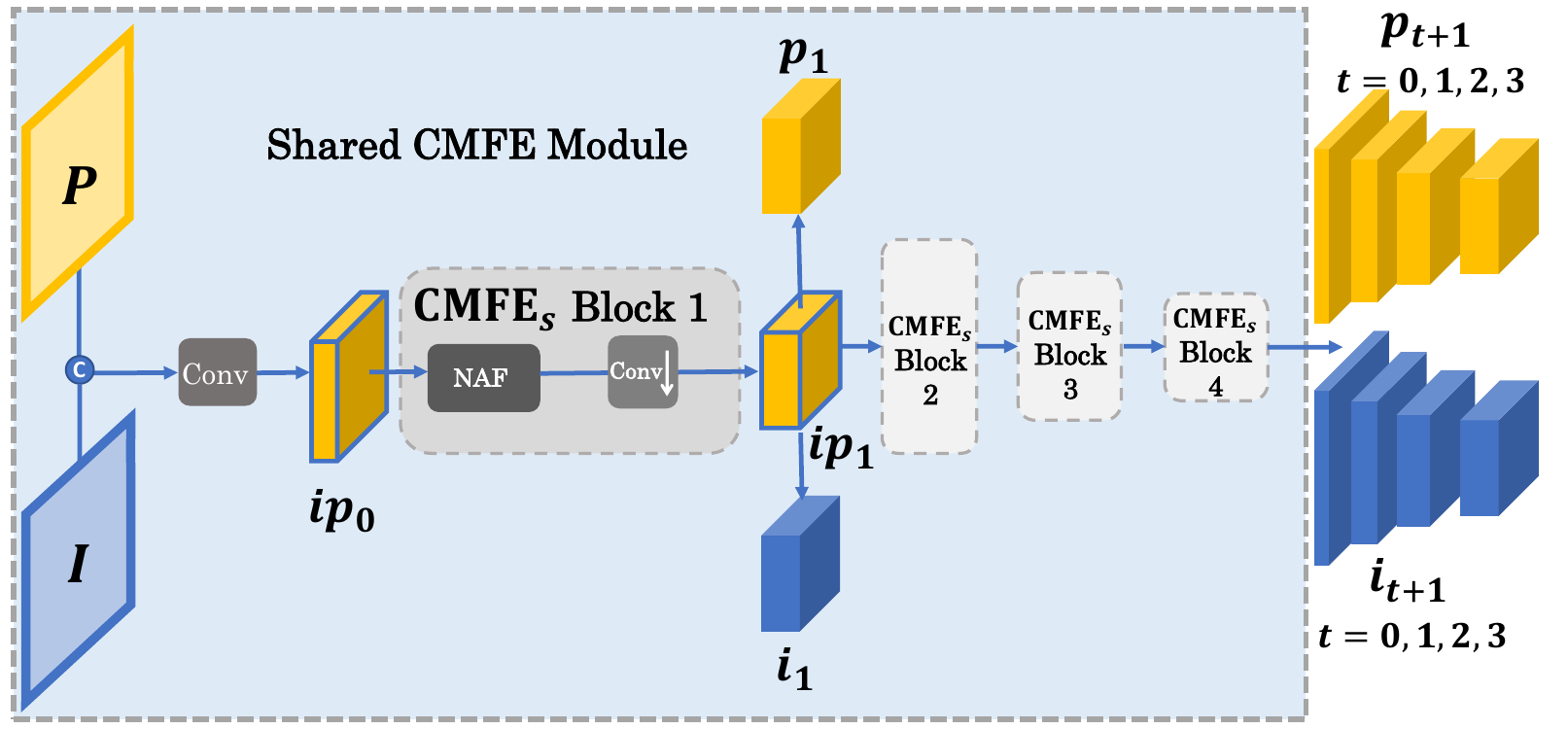}
	\caption{The detailed structure of the CMFE$_s$.}
	\label{fig:cmfe_s}
\end{figure}

\begin{figure*}[t]
	\setlength{\abovecaptionskip}{0.2cm}
	\centering
	\includegraphics[width=1\linewidth]{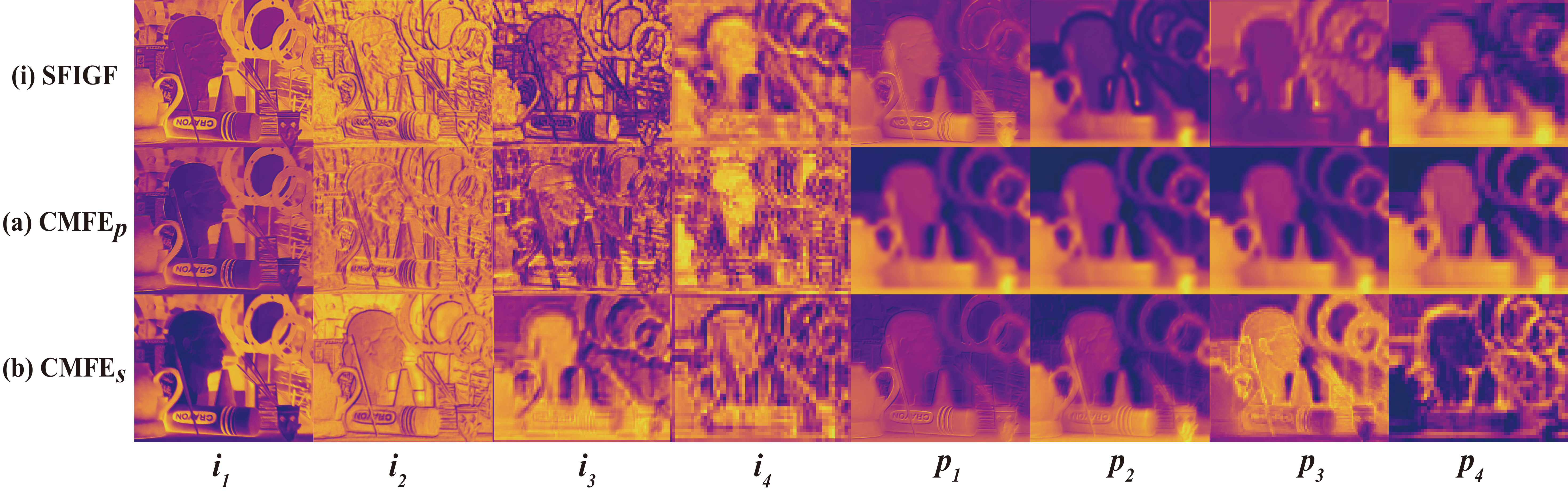}
	\caption{ Visualization of the extracted multi-scaled features by the different variants of the CMFE.}
	\label{fig:cmfe}
\end{figure*}

In the main text, we have experimented with two variants of the CMFE module, i.e., CMFE$_p$ and CMFE$_s$. Here, we first show the detailed structures of these two variants as in Fig.~\ref{fig:cmfe_p} and Fig.~\ref{fig:cmfe_s}, respectively. Then, in addition to the final results shown in the main text, we visualize the extracted features by different variants in Fig.~\ref{fig:cmfe}. It can be observed that, compared with the proposed SFIGF, both the $i_t$s and $p_t$s of CMFE$_{p}$, which are extracted separately, are unable to reflect the necessary structural contents without the mutual information interaction. In contrast, employing the completely shared CMFE$_{s}$ to extract $i_{t}$s and $p_{t}$s can preserve similar structural information across the two representations while also retaining unexpected texture details. 


\subsection{Effectiveness of the FeGF module}
\begin{figure*}[t]
	\setlength{\abovecaptionskip}{0.2cm}
	\centering
	\includegraphics[width=1\linewidth]{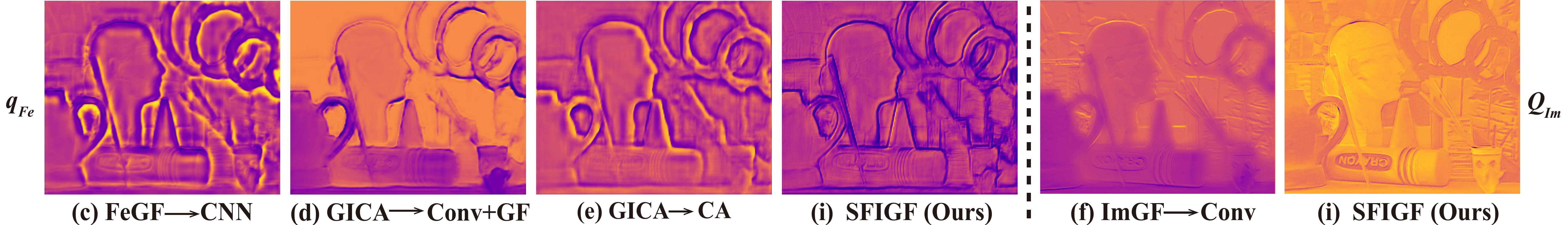}
	\caption{Visualization of the intermediate results in the feature domain (left) and image domain (right) with different variants of modules as introduced in Section 5.5 of the main text.}
	\label{fig:qimqfe}
\end{figure*}

For a deeper insight into the FeGF module, we further visualize the feature domain outputs of all its variants considered in the main text, and the following observations can be drawn.
First, as shown in Fig.~\ref{fig:qimqfe} (c), replacing the whole FeGF module with convolutional blocks leads to a loss of object structures at the feature level. 
Second, the feature output shown in Fig.~\ref{fig:qimqfe} (d) contains more sharp edges, which can be attributed to the introduction of the GF mechanisms, but still performs worse than the complete SFIGF as shown in Fig.~\ref{fig:qimqfe} (i). Third, leveraging the long-range dependency by CA, the feature restoration depicted in Fig.~\ref{fig:qimqfe} (e) exhibits a more favorable performance compared with both Fig.~\ref{fig:qimqfe} (c) and Fig.~\ref{fig:qimqfe} (d), but the absence of the full GF mechanism results in a worse edge restoration with ghosting, as compared with Fig.~\ref{fig:qimqfe} (i).

\subsection{Effectiveness of the ImGF module}

We replace the proposed ImGF module with convolutional layers, where the output $Q_{Im}$ can be achieved by:
\begin{gather}
	Q_{Im} = \mathrm{Conv}(\mathrm{Cat}(i_1, p_1, I,P)),
\end{gather}
where $\mathrm{Conv}(\cdot)$ refers to convolution blocks with activation function GELU, and $\mathrm{Cat}(\cdot)$ refers to the concatenation operation.
The image domain restoration result of this variant is shown in Fig.~\ref{fig:qimqfe} (f). It can be seen that without the instruction of the GF mechanism, the result suffers from degradation in detail restoration. This leads to the unsatisfactory final output shown in the main text.     

\begin{figure}[t]
	\setlength{\abovecaptionskip}{0.2cm}
	\centering
	\includegraphics[width=0.9\linewidth]{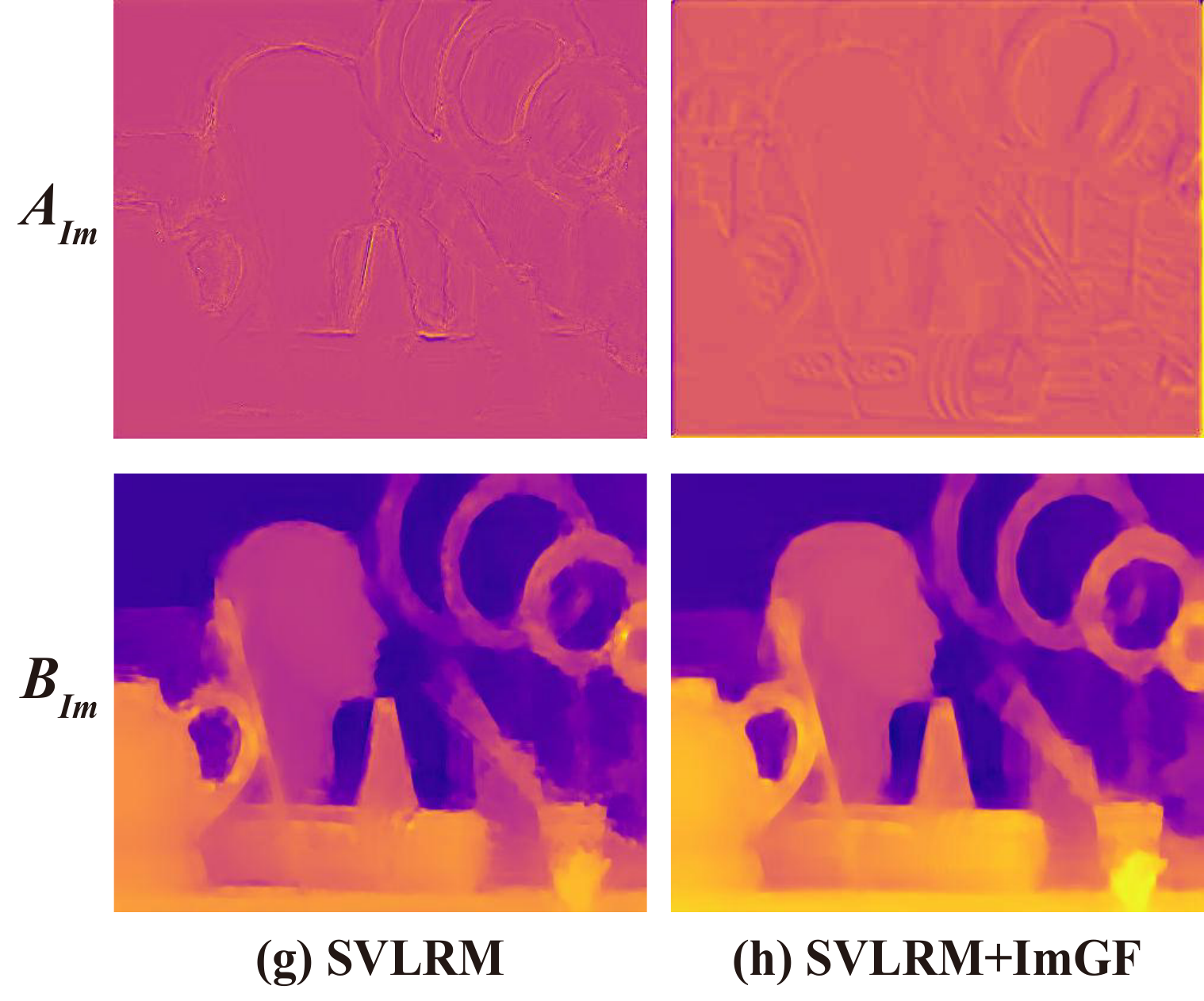}
	\caption{Comparison of the learned $A_{Im}$ and $B_{Im}$ by SVLRM (left) and its improvement by our ImGF mechanism (right).}
	\label{fig:sv_im}
\end{figure}

We also present additional feature visualization of SVLRM and SVLRM with ImGF in Fig.~\ref{fig:sv_im}. It can be seen that our proposed ImGF operation can help to achieve $A_{Im}$ with more details and $B_{Im}$ with more even surface and clearer edges, which can be attributed to that it more closely follows the GF mechanism.


\section{Detailed experimental settings and more visual results}
\label{sec:sup_exp}
In this part, We provide detailed experimental settings and more visual results for all the 4 GIR tasks considered in Section 5 of the main text. It should be noted that we apply different sizes of the SFIGF for different tasks, which can be controlled by setting a proper number of the base channels ($n$ in Table~\ref{tab:ch}).

\subsection{GDSR experiments}
\subsubsection{Detailed experiment settings}
The GDSR task aims at restoring an HR depth map from an LR one under the guidance of an RGB image captured in the same scene. The LR depth map $P \in \mathbb{R}^{h\times w \times 1}$ is the to-be-restored image, and the HR RGB image $I \in \mathbb{R}^{H\times W \times 3} $ is regarded as the guidance, where $h<H$ and $w<W$. For many deep learning methods, $P$ is generally pre-upsampled, e.g., by bicubic interpolation, to be with the same spatial size of $I$, and we follow this preprocessing strategy. Then the output should be a reconstructed depth map $Q_{out}\in \mathbb{R}^{H\times W \times 1}$. In this task, we train SFIGF in a supervised way with $L_1$ Loss for 200 epochs, and the initial learning rate is set to $1\times 10^{-4}$, which decays by multiplying a factor of 0.2 every 60 epochs. The batch size is 1, and the patch size is $256\times256$. The number of base channels, $n$, is set to 48 in this task.
Except for DCTNet \cite{zhao2022dct} and SSDNet \cite{zhao2023spherical}, whose pre-trained models are released by authors, we retrain other competing methods under the same settings.
\subsubsection{Visual results}
\begin{figure*}[t]
	\setlength{\abovecaptionskip}{0.2cm}
	\centering
	\includegraphics[width=1\linewidth]{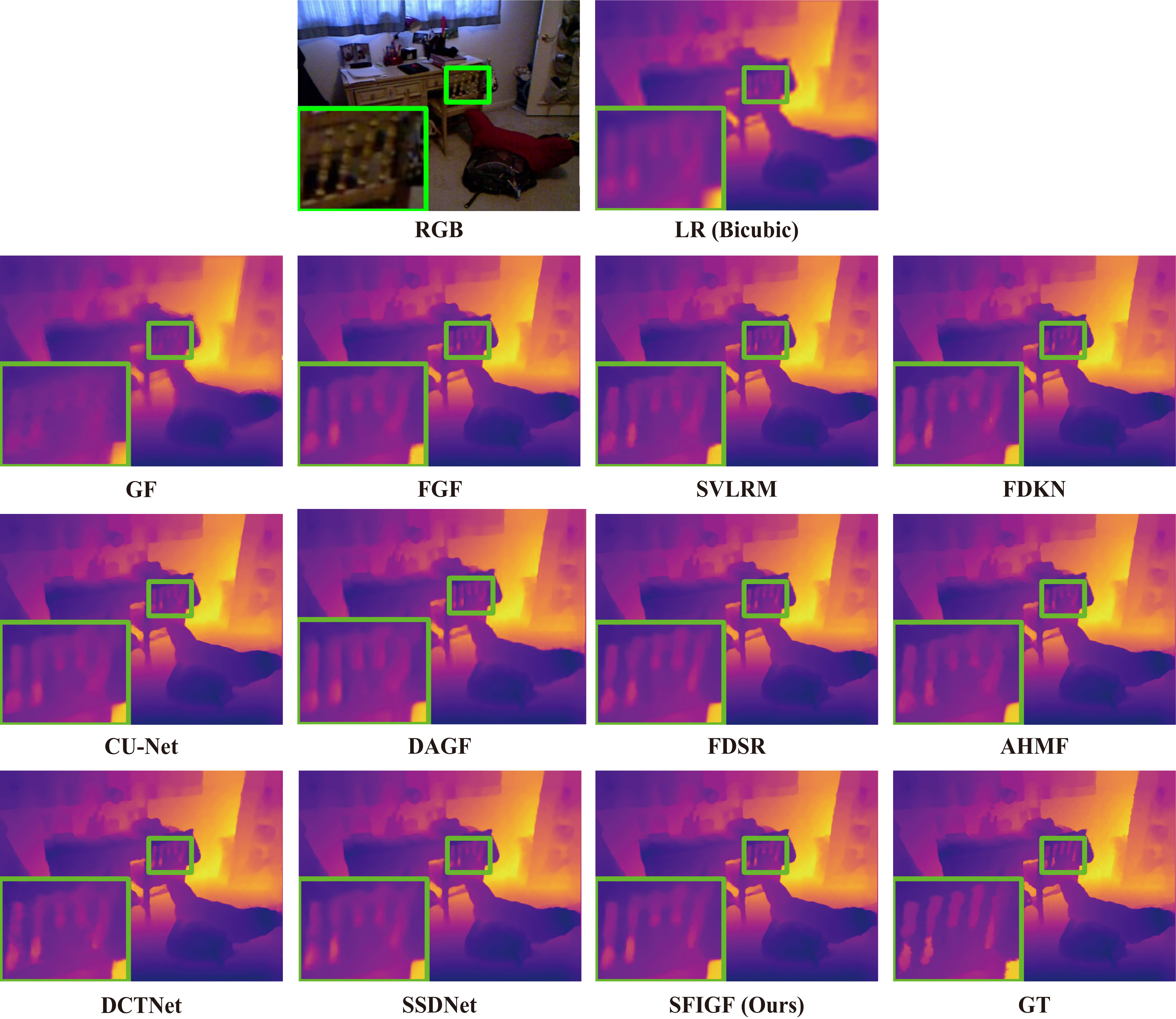}
	\caption{ Visual results of 8x GDSR on NYU v2.}
	\label{fig:nyu_0}
\end{figure*}

\begin{figure*}[t]
	\setlength{\abovecaptionskip}{0.2cm}
	\centering
	\includegraphics[width=1\linewidth]{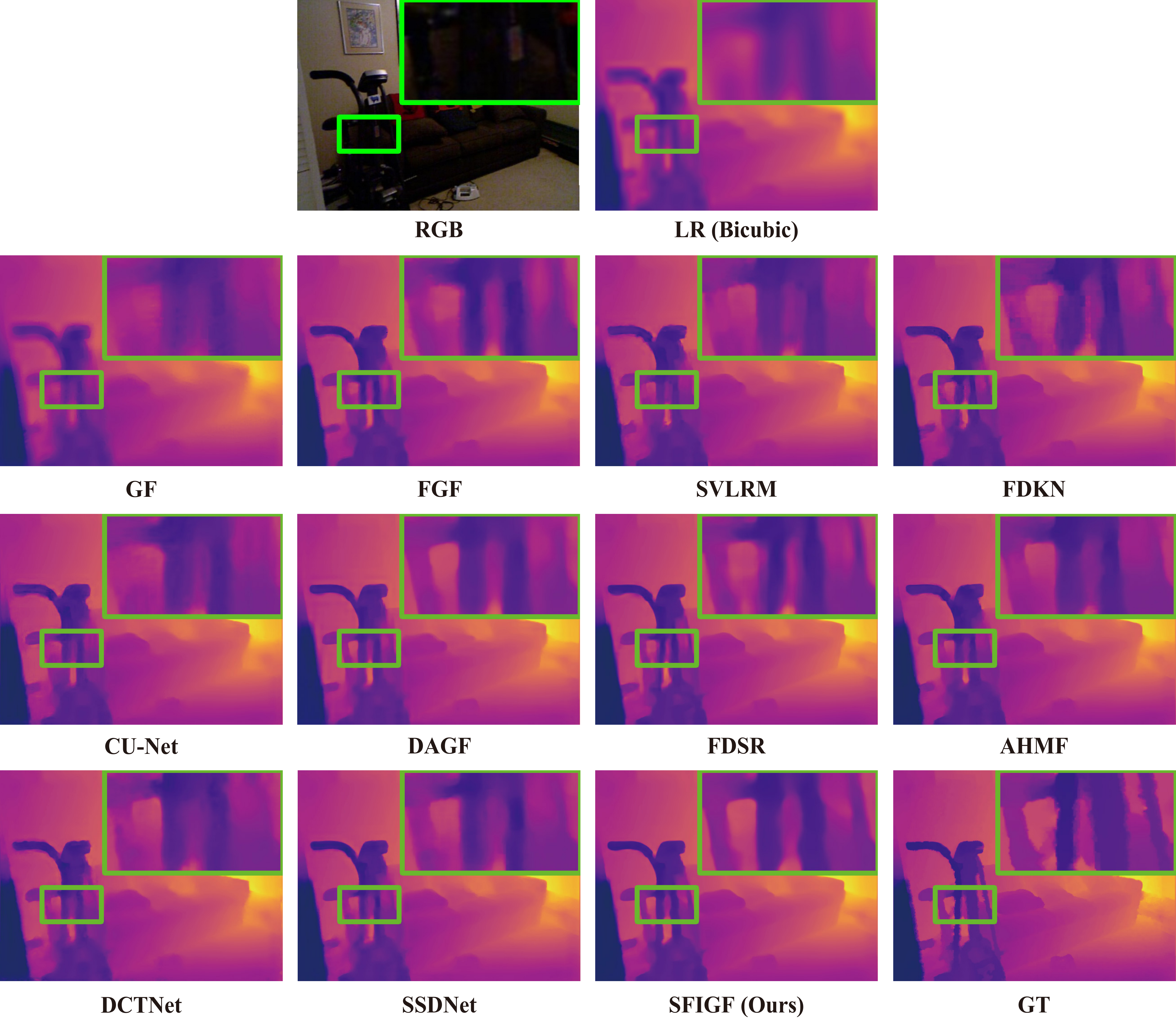}
	\caption{ Visual results of 16x GDSR on NYU v2.}
	\label{fig:nyu_1}
\end{figure*}

\begin{figure*}[t]
	\setlength{\abovecaptionskip}{0.2cm}
	\centering
	\includegraphics[width=1\linewidth]{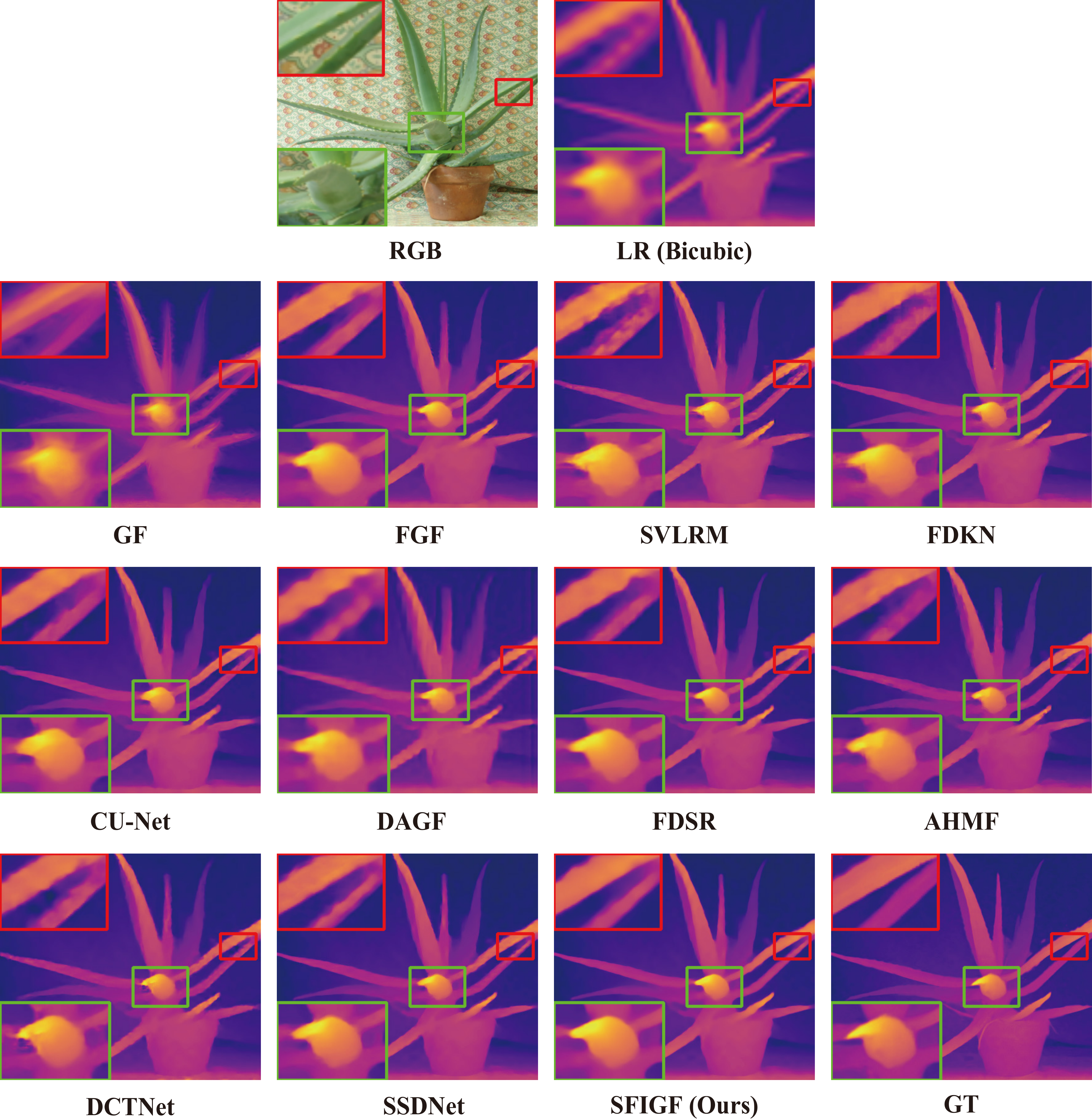}
	\caption{ Visual results of 8x GDSR on Middlebury.}
	\label{fig:mid_0}
\end{figure*}

\begin{figure*}[t]
	\setlength{\abovecaptionskip}{0.2cm}
	\centering
	\includegraphics[width=1\linewidth]{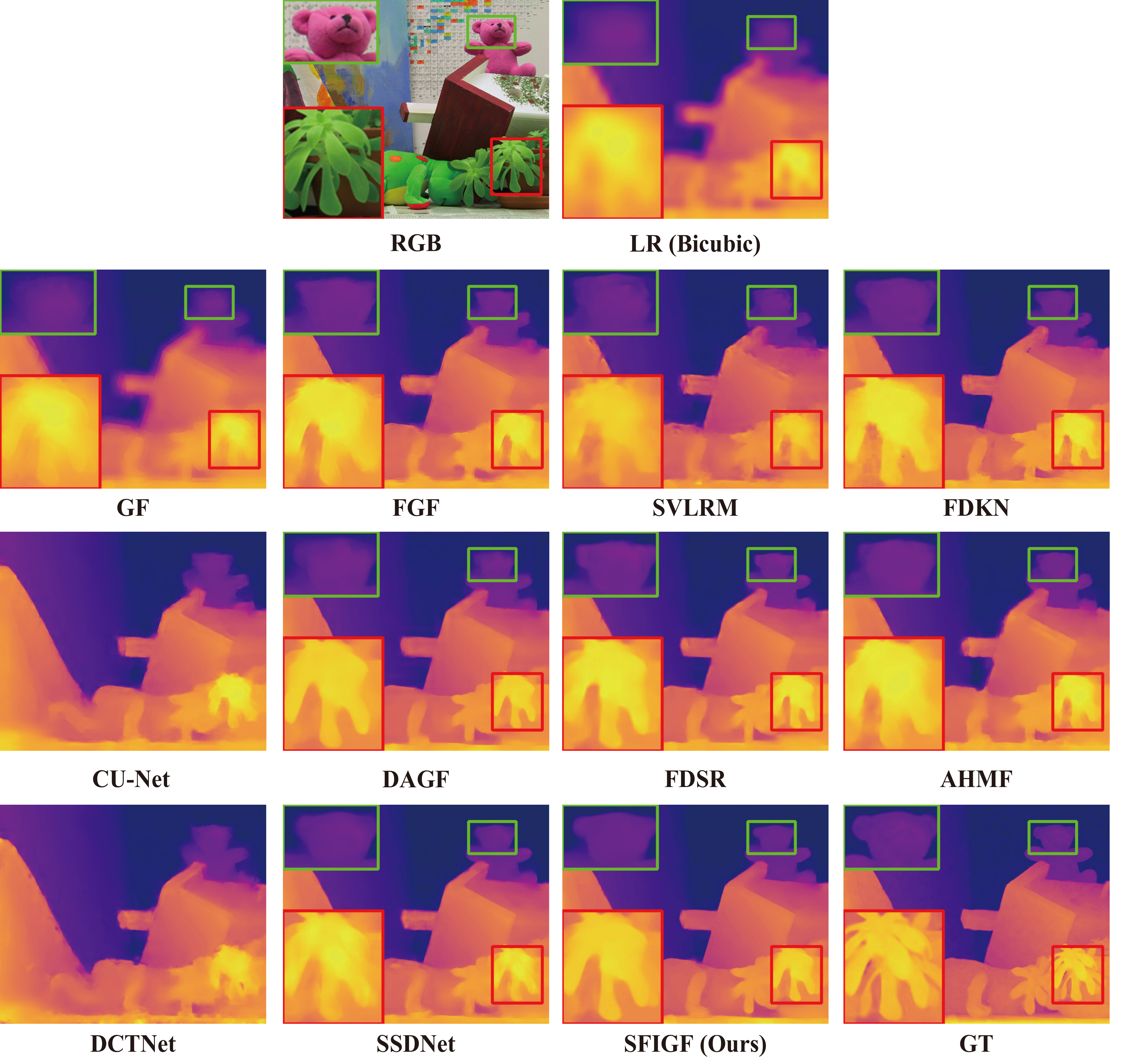}
	\caption{ Visual results of 16x GDSR on Middlebury.}
	\label{fig:mid_1}
\end{figure*}

\begin{figure*}[t]
	\setlength{\abovecaptionskip}{0.2cm}
	\centering
	\includegraphics[width=1\linewidth]{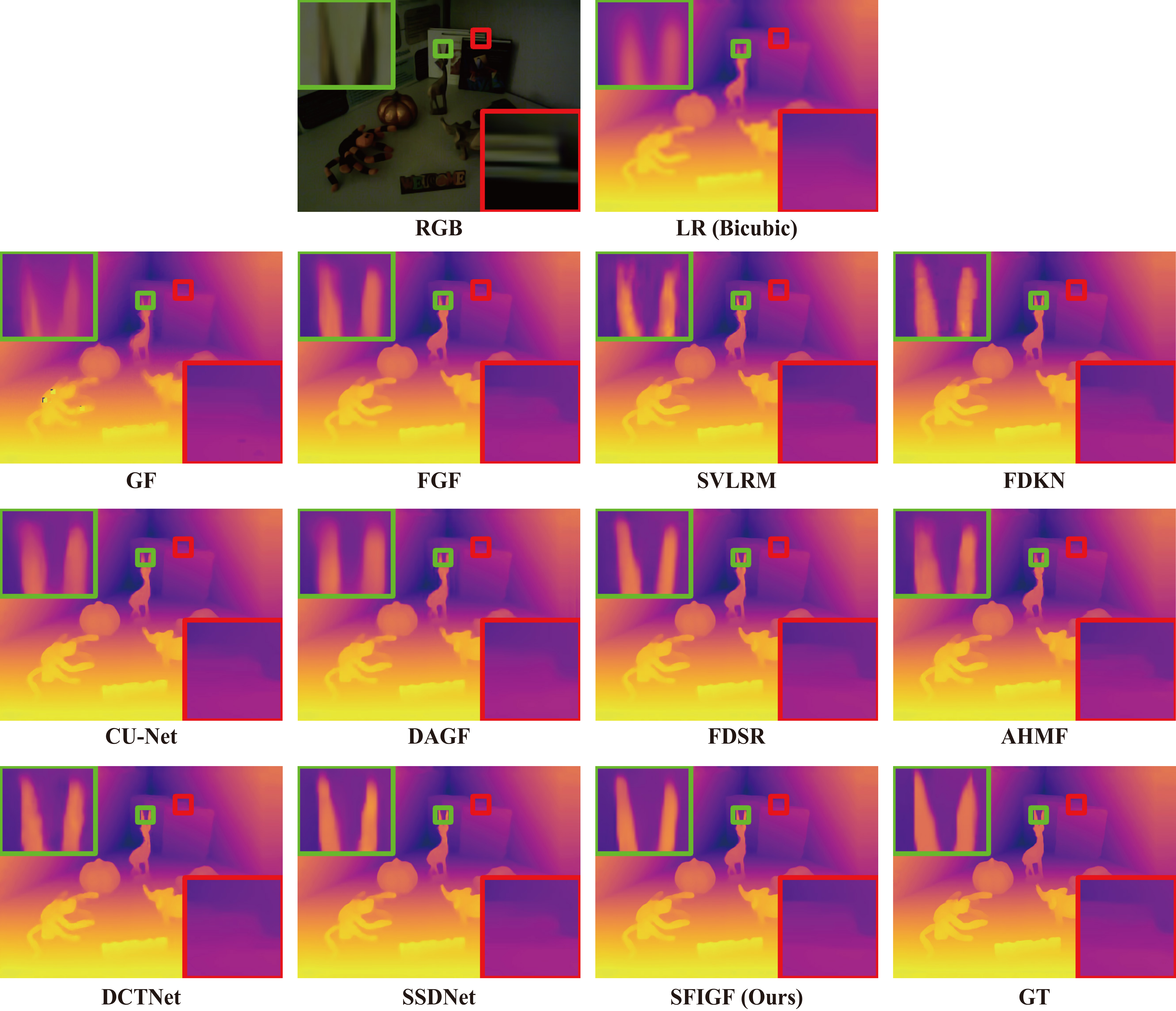}
	\caption{ Visual results of 8x GDSR on Lu.}
	\label{fig:lu_0}
\end{figure*}

\begin{figure*}[t]
	\setlength{\abovecaptionskip}{0.2cm}
	\centering
	\includegraphics[width=1\linewidth]{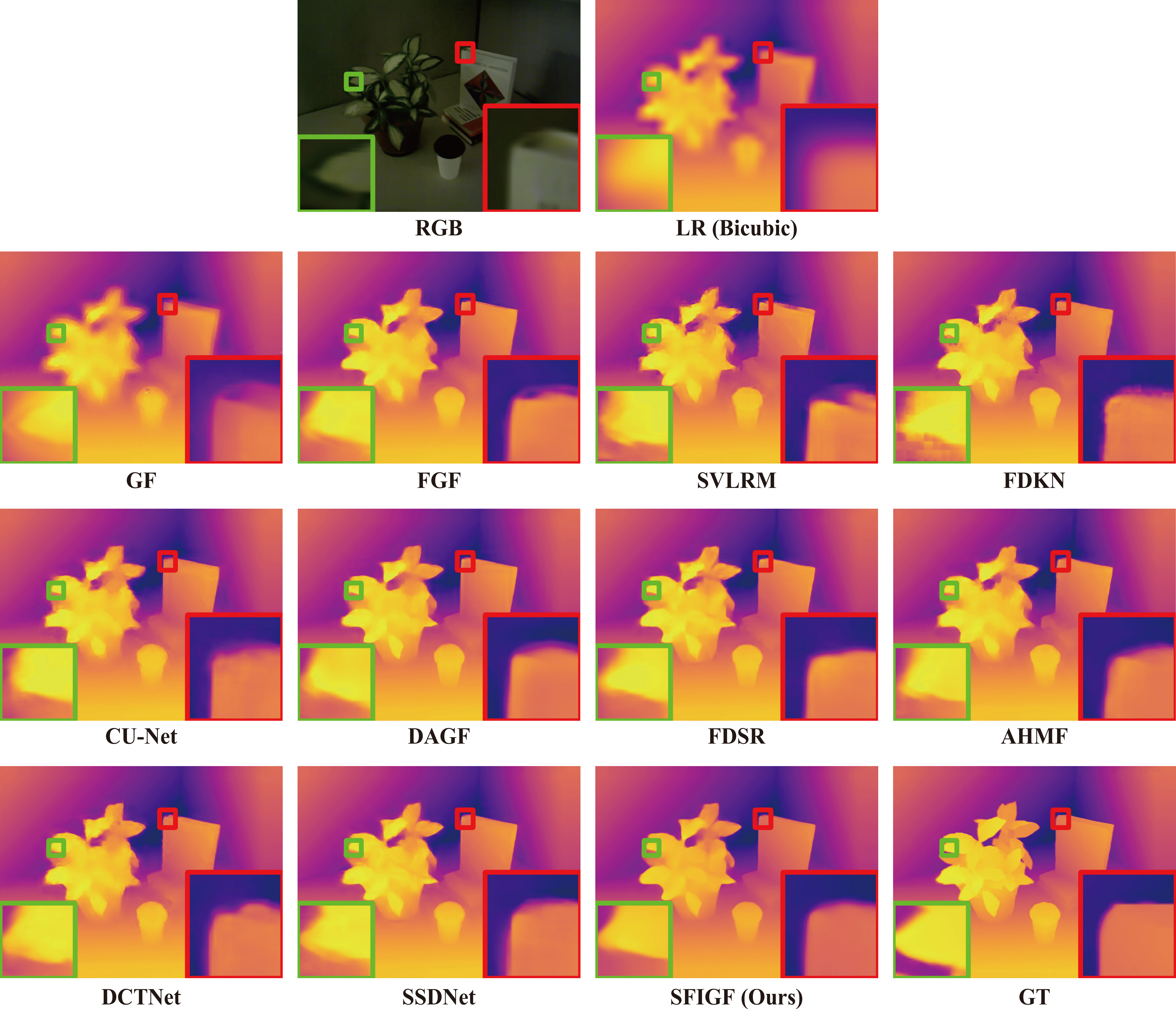}
	\caption{ Visual results of 16x GDSR on Lu.}
	\label{fig:lu_1}
\end{figure*}

We provide here more visual results of 8x and 16x GDSR on the NYU v2 \cite{silberman2012indoor}, Middlebury \cite{hirschmuller2007mid}, and Lu \cite{lu2014depth} datasets, as shown in Figs.~\ref{fig:nyu_0}-\ref{fig:lu_1}.
It can be clearly seen from the results that our method outperforms both the general-purpose GIR methods and the task-specified deep models.


\subsection{Pan-sharpening experiments}
\subsubsection{Detailed experimental settings}
As mentioned in the main text, the pan-sharpening problem focuses on restoring an HRMS image $Q_{out} \in \mathbb{R} ^ {h\times w \times C}$ from an LRMS image $P \in \mathbb{R} ^ {H\times W \times C}$ with the help of an HR panchromatic image $I \in \mathbb{R} ^ {H\times W \times 1}$, where $h<H$ and $w<W$ and the channel number $C$ is varied according to datasets, which is 8 in our experiments. Similar to GDSR, the LRMS image $P$ can first be pre-upsampled with spatial size $H\times W$, and the upsample scale is 4. In the full-resolution WorldView-\uppercase\expandafter{\romannumeral3} dataset, the spatial height $H$ and width $W$ are both 256. 
The training and testing datasets are pre-split.
We train our SFIGF in a supervised way with $L_1$ Loss for 3000 epochs. The batch size is 10, and the patch size is $128\times 128$. The initial learning rate is $1\times 10^{-4}$ and decreases as the training iterations increase. The number of base channels is set to 32. All methods are retrained under the same setting. 
It should be mentioned that, as demonstrated by Deng et al. \cite{jin2022lagconv}, the introduction of multi-scale structures does not improve performance for the pan-sharpening task but tends to lead to a loss of details because of the downsampling operations. Therefore, we only use the one-scale structure for the CMFE and FeGF modules in our SFIGF for this task.
\subsubsection{Visual results}
\begin{figure*}[t]
	\setlength{\abovecaptionskip}{0.2cm}
	\centering
	\includegraphics[width=1\linewidth]{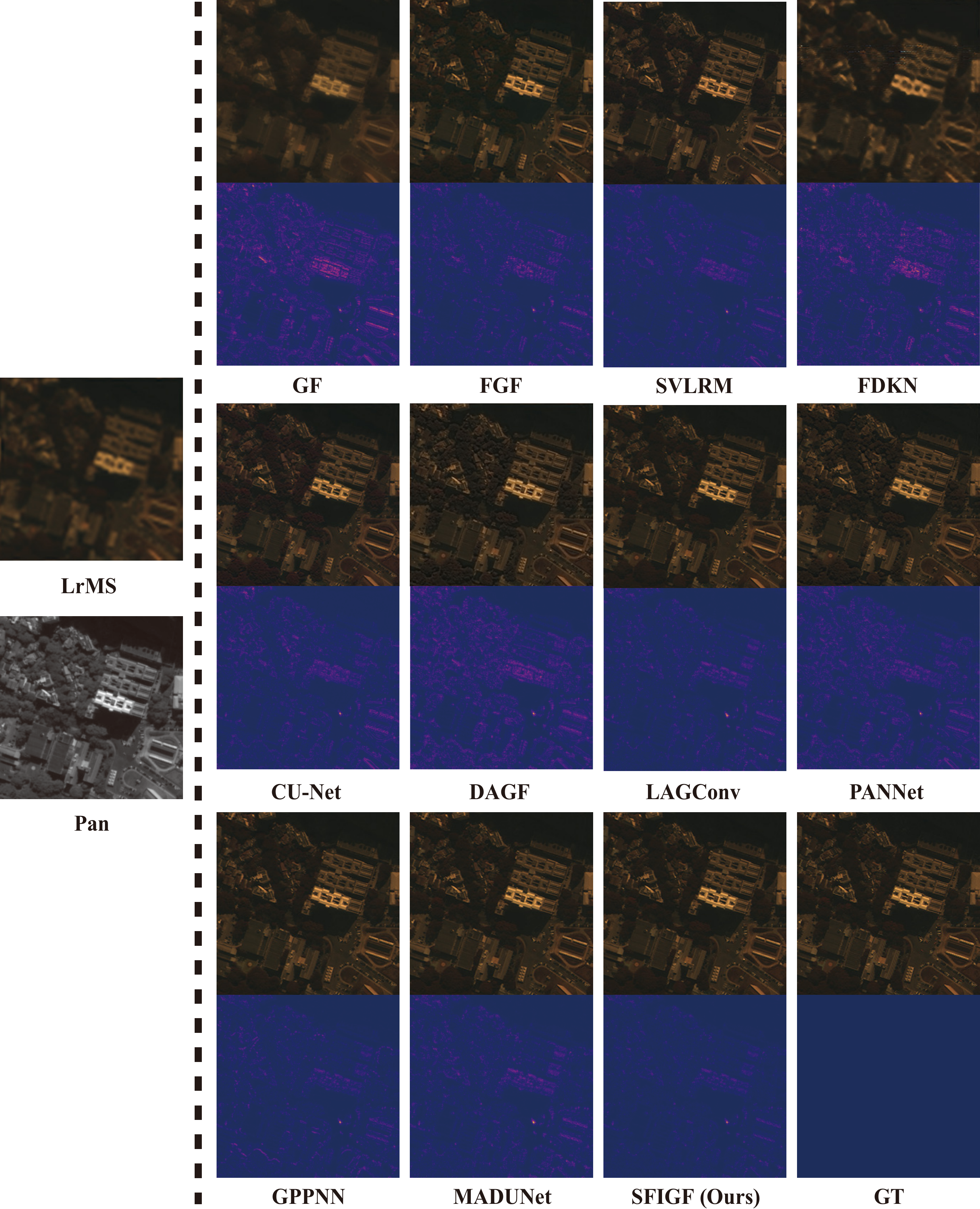}
	\caption{ Visual results of competed methods on the WorldView-\uppercase\expandafter{\romannumeral3} of the pan-sharpening task.}
	\label{fig:pan_0}
	\vspace{-0.5cm}
\end{figure*}

\begin{figure*}[t]
	\setlength{\abovecaptionskip}{0.2cm}
	\centering
	\includegraphics[width=1\linewidth]{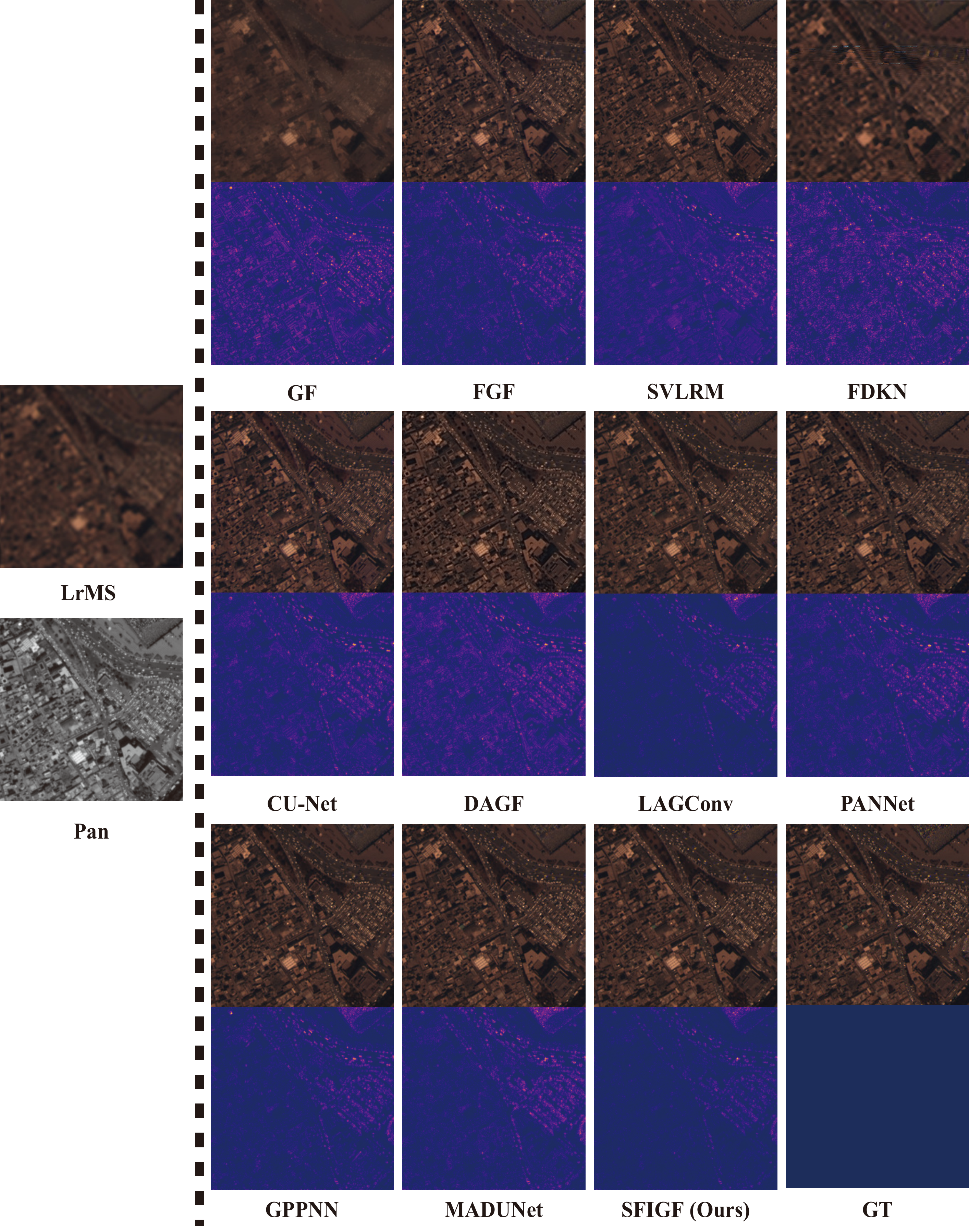}
	\caption{ Visual results of competed methods on the WorldView-\uppercase\expandafter{\romannumeral3} of the pan-sharpening task.}
	\label{fig:pan_1}
	\vspace{-0.5cm}
\end{figure*}

The visual results are shown in Figs.~\ref{fig:pan_0}-\ref{fig:pan_1}. The top row presents the restored HRMS results, while the bottom row displays the corresponding RMSE map. Evidently, our SFIGF produces fewer residuals and outperforms both the general-purpose GIR methods and the specialized pan-sharpening methods, while exhibiting richer information on structures.



\subsection{MFIF experiments}
\subsubsection{Detailed experimental settings}
For the MFIF task, we employ a dual-guidance framework. Specifically, we use two SFIGFs as two branches and let the two input images guide each other. Then the mid-outputs of the two SFIGF branches are fused at the output layer to generate the final result. The whole process can be expressed as follows:
\begin{gather}
    Q_{out}^1 = \mathrm{SFIGF}_1(I_1, I_2),\\
    Q_{out}^2 = \mathrm{SFIGF}_2(I_2, I_1),\\
    Q_{out} = \mathrm{Conv}(\mathrm{Cat}[Q_{out}^1, Q_{out}^2])
\end{gather}
where $I_1, I_2 \in \mathbb{R}^{H\times W\times3}$ refers to input images captured with different focal distances.
Besides, as mentioned in the main text, the MFIF task is commonly formulated in an unsupervised manner to meet practical requirements, and thus we utilize two unsupervised losses for network training without relying on ground truth images:
\begin{gather}
    L_2 = s_1\|Q_{out}-I_1\|_2^2 + s_{2}\|Q_{out}-I_2\|_2^2, \label{eq:mffl_1}\\
    L_{grad} = s_{1}\|\nabla Q_{out}-\nabla I_1\|_2^2 + s_{2}\|\nabla Q_{out}-\nabla I_2\|_2^2, \label{eq:mffl_2}
\end{gather}
where $s_{1} = \mathrm{sign}(HF(I_1)-\min(HF(I_1), HF(I_2)))$, $s_{2} = 1 - s_{1}$, $\nabla$ refers to the gradient operator, and $HF(\cdot)$ refers to the high-frequency contents of the image, calculated by the Gaussian filtering kernel. Loss $L_2$ defined in Eq.~\eqref{eq:mffl_1} aims to preserve the content information from near-focus and far-focus inputs simultaneously in the image domain, and $L_{grad}$ defined in Eq.~\eqref{eq:mffl_2} aims to preserve their structural information extracted by gradient operations. The final loss is the combination of two losses:
\begin{equation}
	L = L_2 + L_{grad},
\end{equation}
and such a combination of losses from image and gradient domains has been shown effective in previous studies \cite{ma2022swinfusion, ma2021smfuse}. 

In the MFIF experiments, we train SFIGF for 200 epochs with batch size 16, patch size $128\times128$, and the initial learning rate $1\times10^{-4}$. We also use only one-scale SFIGF as in pan-sharpening experiments, and the number of base channels is set to 32. Except for ZMFF \cite{hu2023zmff}, all methods are retrained under the same setting.

\subsubsection{Visual results}
\begin{figure*}[t]
	\setlength{\abovecaptionskip}{0.2cm}
	\centering
	\includegraphics[width=1\linewidth]{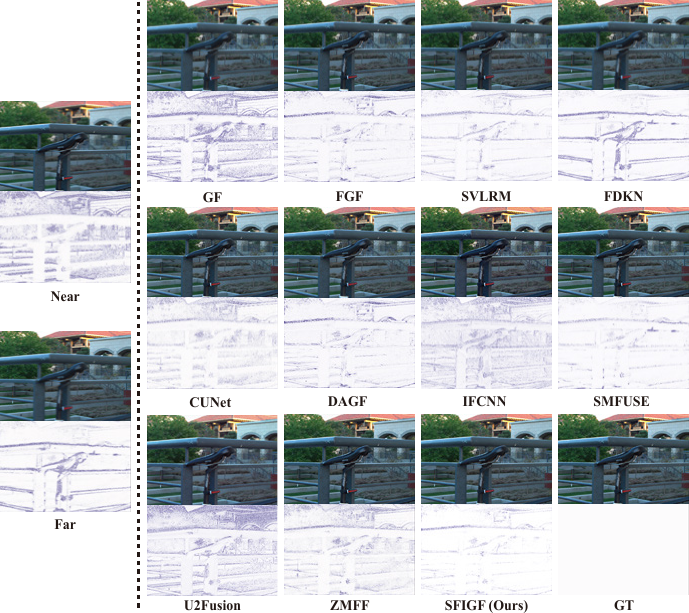}
	\caption{Visual comparison of competed methods on the Real-MFF dataset of the MFIF task. The bottom row refers to the corresponding error maps.}
	\label{fig:mff_0}
\end{figure*}

\begin{figure*}[t]
	\setlength{\abovecaptionskip}{0.2cm}
	\centering
	\includegraphics[width=1\linewidth]{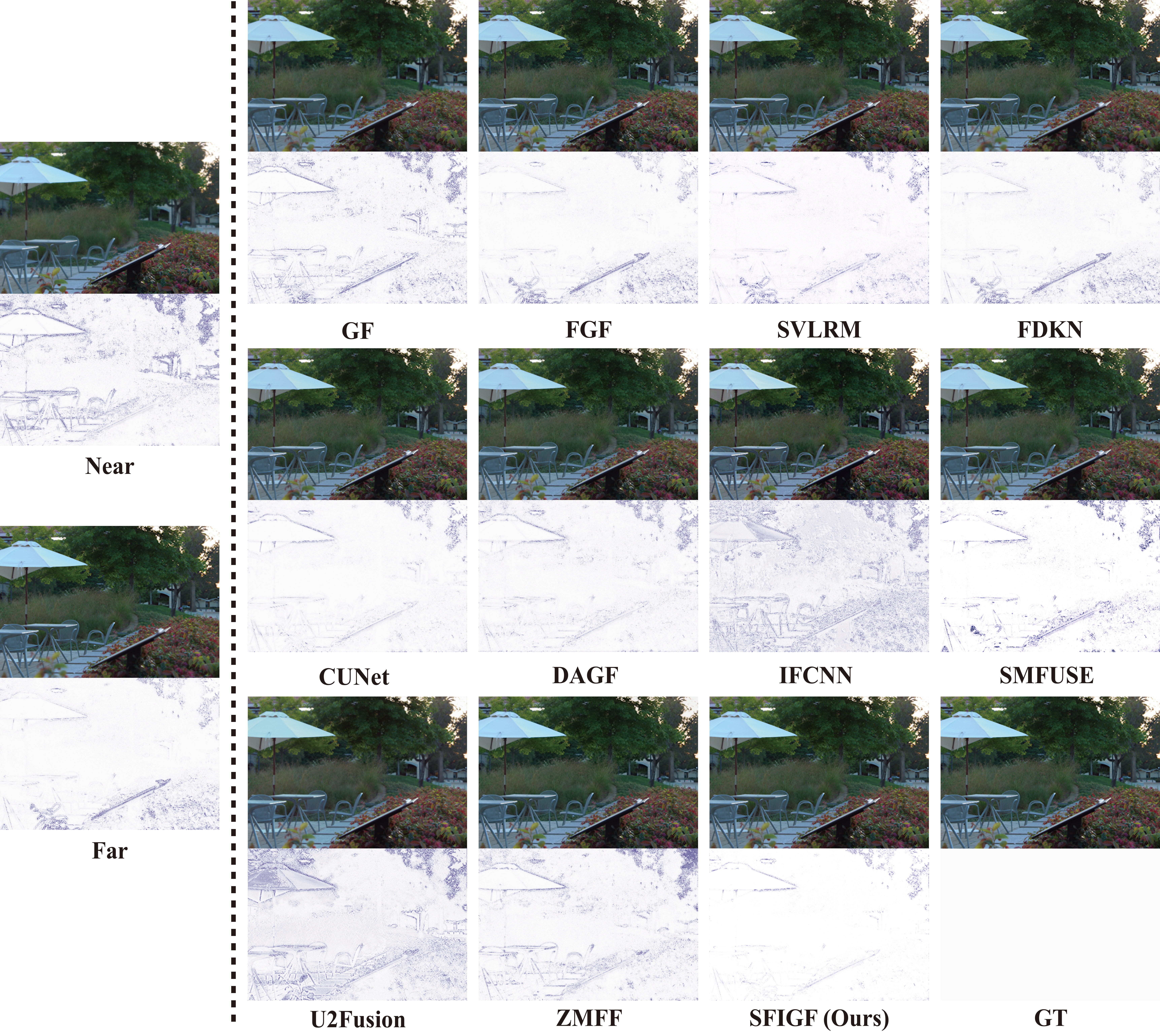}
	\caption{Visual comparison of competed methods on the Real-MFF dataset of the MFIF task. The bottom row refers to the corresponding error maps.}
	\label{fig:mff_1}
\end{figure*}

\begin{figure*}[t]
	\setlength{\abovecaptionskip}{0.2cm}
	\centering
	\includegraphics[width=1\linewidth]{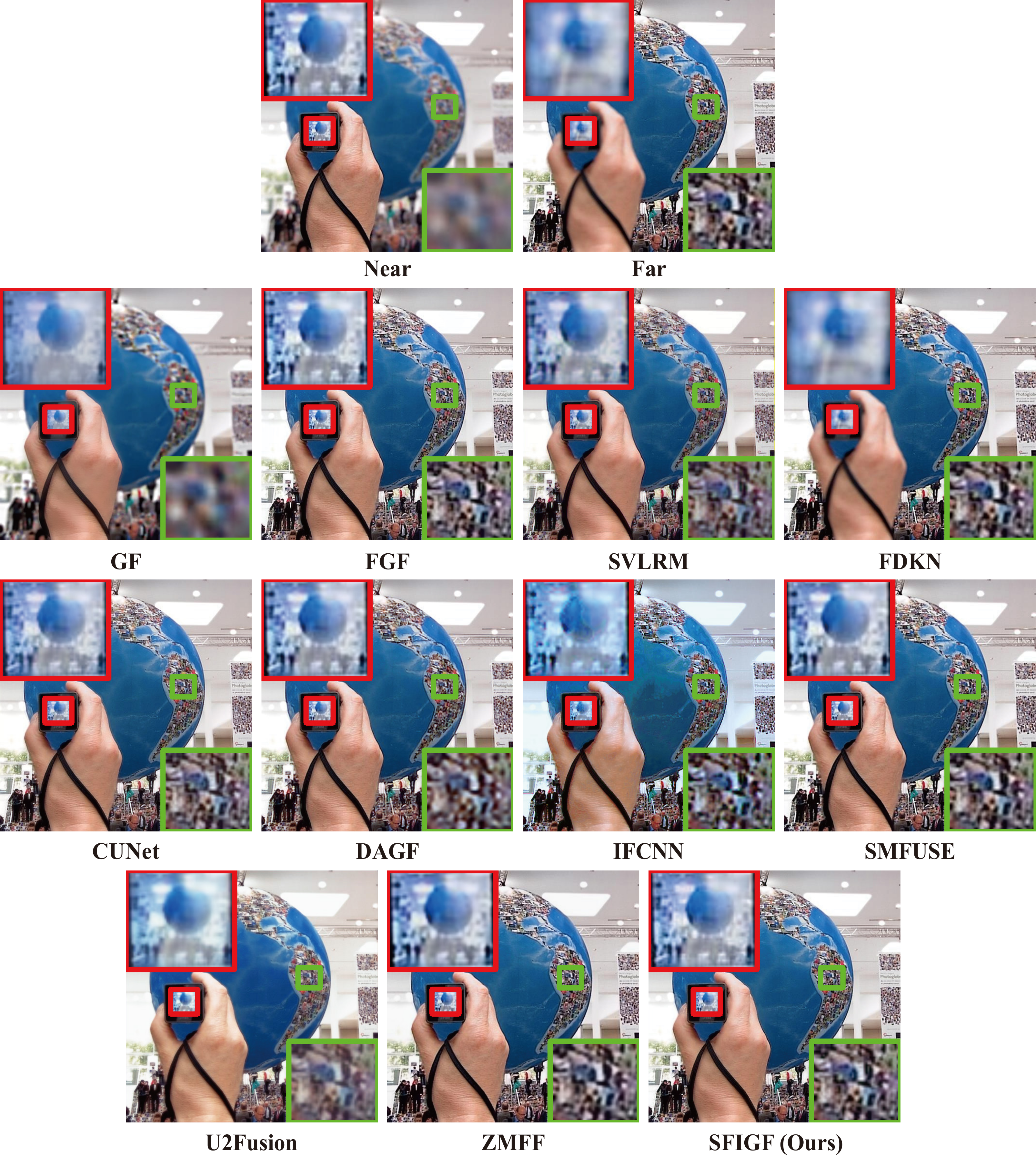}
	\caption{Visual comparison of competed methods on the Lytro dataset of the MFIF task. The bottom row refers to the corresponding error maps.}
	\label{fig:lytro_0}
\end{figure*}

\begin{figure*}[t]
	\setlength{\abovecaptionskip}{0.2cm}
	\centering
	\includegraphics[width=1\linewidth]{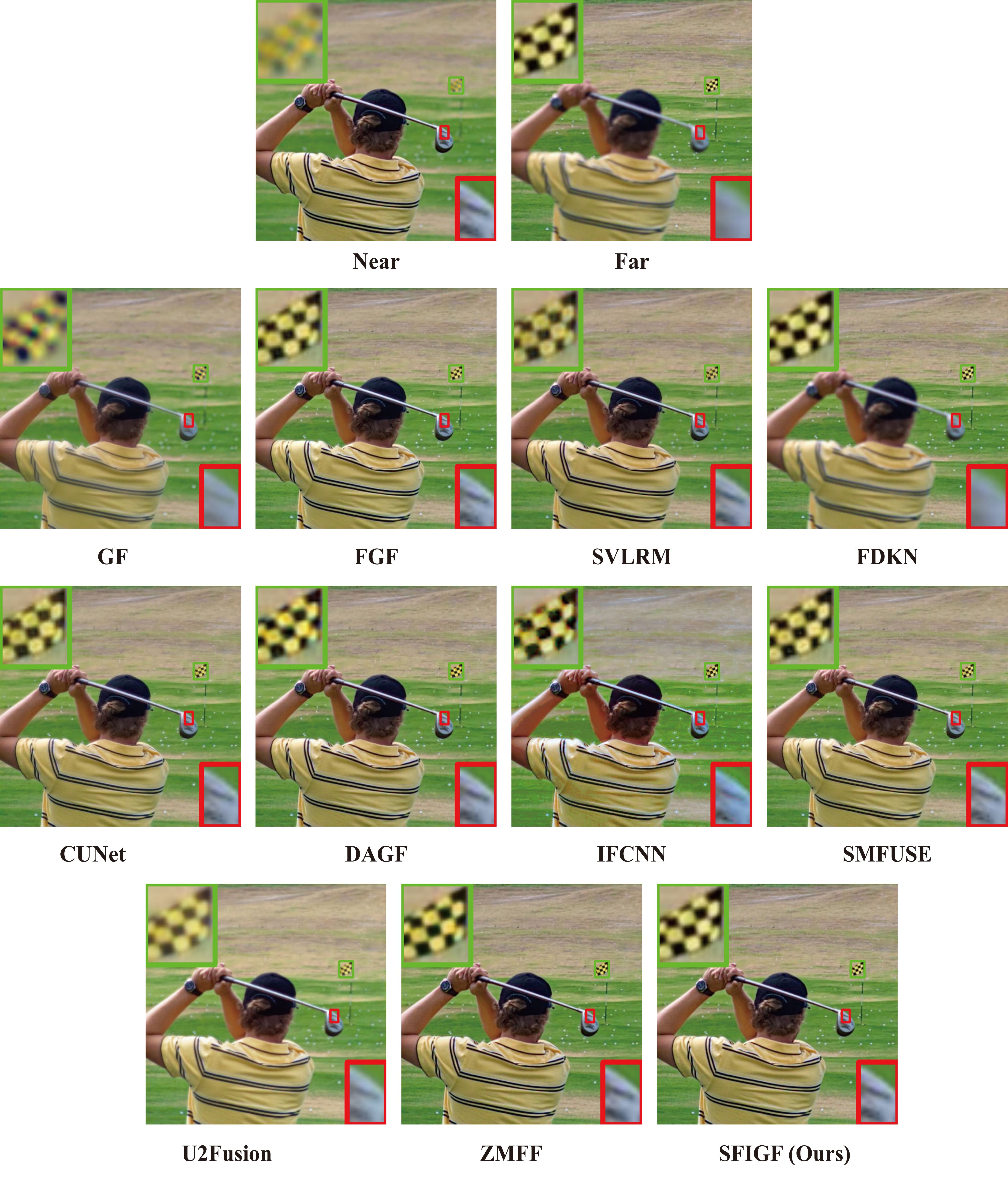}
	\caption{Visual comparison of competed methods on the Lytro dataset of the MFIF task. The bottom row refers to the corresponding error maps.}
	\label{fig:lytro_1}
\end{figure*}

Visual results on the RealMFF and Lytro datasets are shown in Figs.~\ref{fig:mff_0}-\ref{fig:mff_1}, and Figs.~\ref{fig:lytro_0}-\ref{fig:lytro_1}, respectively. For the RealMFF dataset, which has ground-truth images, we present the fused results in the top row and the corresponding error map in the bottom row, following \cite{hu2023zmff}.
These visual results on both datasets further verify the superiority of the proposed SFIGF over existing methods by better fusing complementary information from near-focus and far-focus images.

\subsection{Guided LRIE experiments}
\subsubsection{Detailed experimental settings}
As introduced in the main text, we consider two scenarios of guided LRIE tasks: (1) RAW image with red flashlight guidance and (2) RAW image with monochrome guidance. 

For the first scenario, we synthesize paired RAW/red flashlight dataset from the Sony subset of the SID dataset \cite{chen2018learning}, which is constructed by extremely low-light RAW images. Images in the Sony subset were captured by the Sony $\alpha$7S \uppercase\expandafter{\romannumeral2} Bayer sensor with the size $4256 \times 2848 \times 1$. The red input image is the red channel of the ground-truth image with the size of $4256 \times 2848 \times 1$. The input images are all packed into 4 channels following \cite{chen2018learning}, and the output is in the sRGB format with 3 channels.
We use 185 paired images for training and 50 paired images for testing.
The patch size is $512\times512$ and the batch size is 1. The learning rate is set to $1\times10^{-4}$. All methods are retrained under the same setting.

For the second scenario, we adopt the Mono-Colored RAW (MCR) paired dataset collected by Dong et al. \cite{dong2022abandoning}. 
The training set includes 3600 paired images, and the testing set includes 384 paired images. In the training, each pair includes an input low-light RAW image, a ground-truth monochrome frame, and a ground-truth normal-light sRGB image.
Following \cite{dong2022abandoning}, we first build a U-Net \cite{ronneberger2015u} for generating a monochrome frame with the ground truth and then fuse the generated one with the original RAW input.
The patch size is $512\times512$ and the batch size is 12. The both inputs have 4 channels and the output has 3 channels. The learning rate is set to $1\times10^{-4}$. Except for DBLE \cite{dong2022abandoning}, all methods are retrained under the same setting.

\subsubsection{Visual results}
\begin{figure*}[t]
	\setlength{\abovecaptionskip}{0.2cm}
	\centering
	\includegraphics[width=1\linewidth]{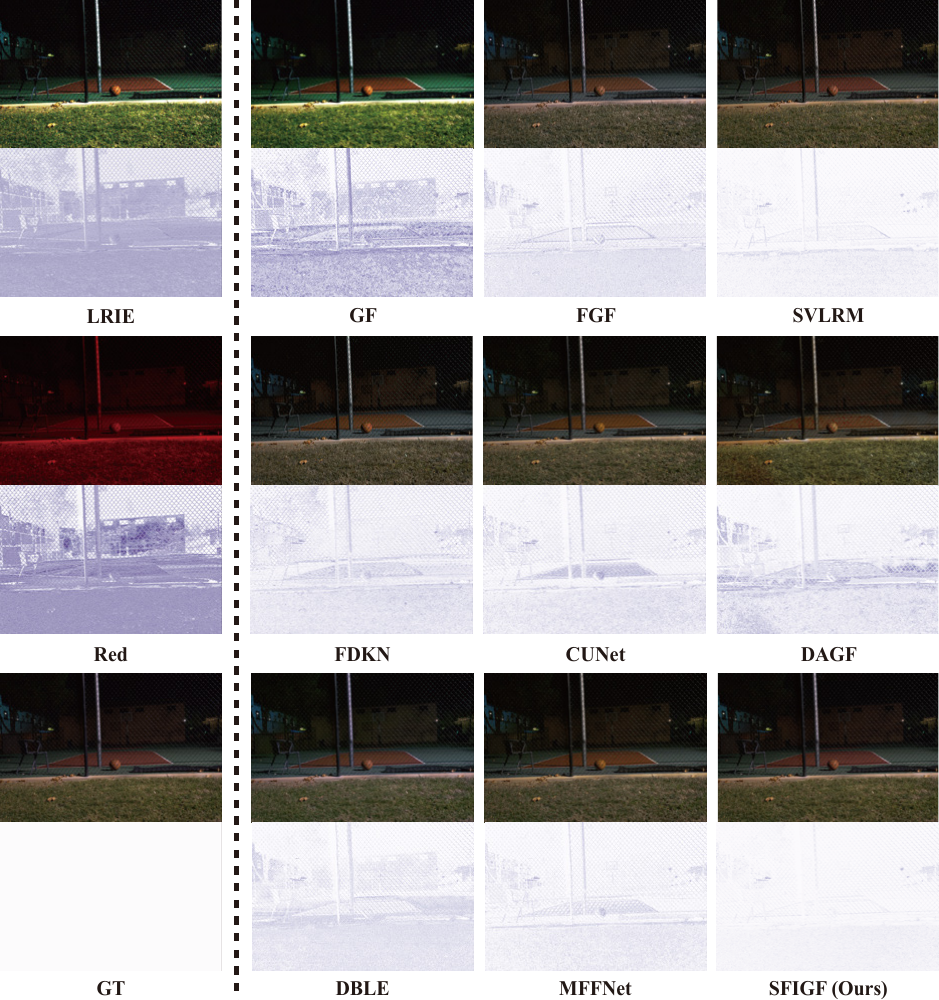}
	\caption{Visual comparison of competed methods on the SID-Sony dataset of the red-guided LRIE task. The bottom row refers to the corresponding error maps.}
	\label{fig:red_0}
\end{figure*}

\begin{figure*}[t]
	\setlength{\abovecaptionskip}{0.2cm}
	\centering
	\includegraphics[width=1\linewidth]{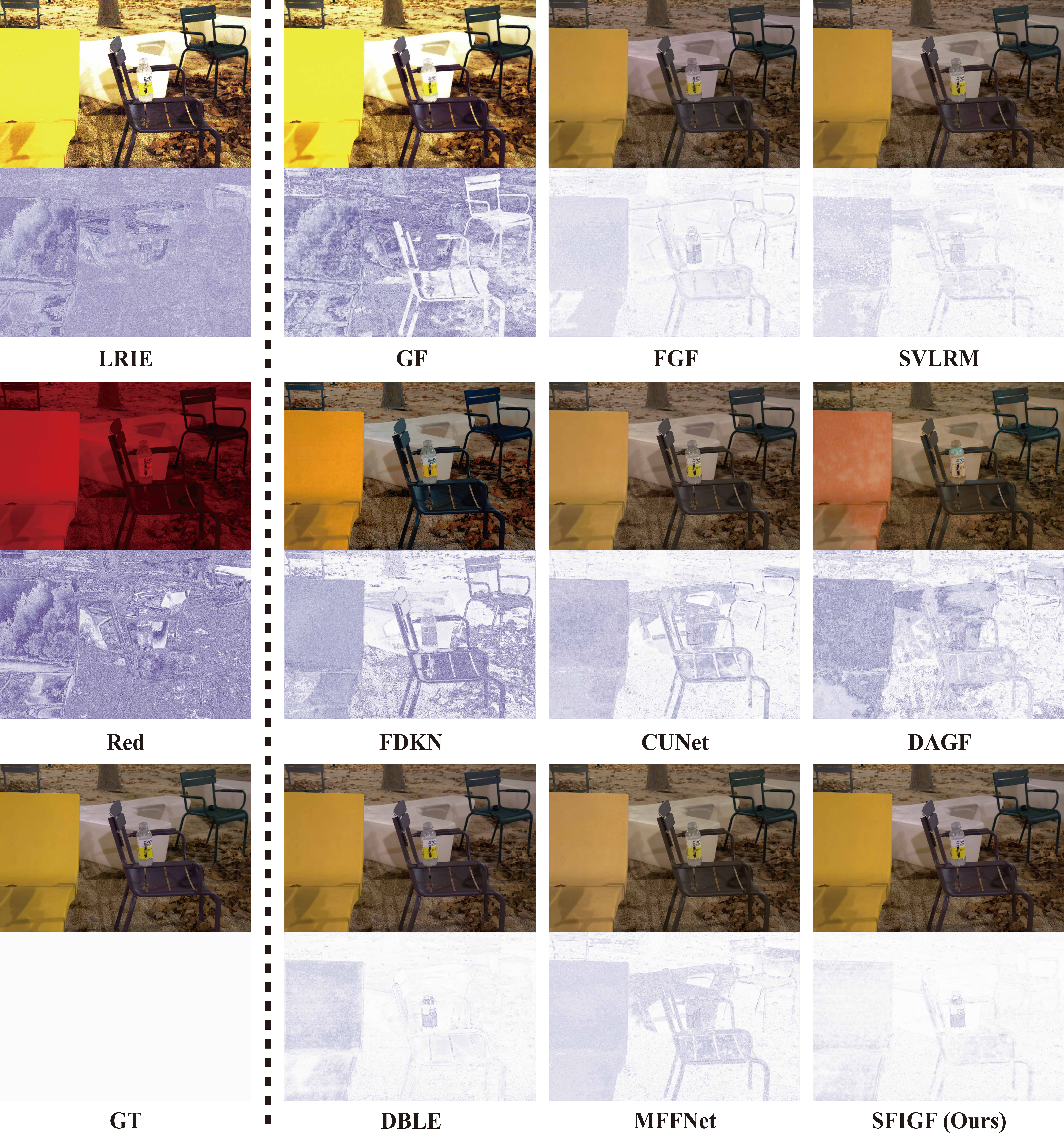}
	\caption{Visual comparison of competed methods on the SID-Sony dataset of the red-guided LRIE task. The bottom row refers to the corresponding error maps.}
	\label{fig:red_1}
\end{figure*}

Visual results on the SID and MCR datasets are shown in Figs.~\ref{fig:red_0}-\ref{fig:red_1}, and Figs.~\ref{fig:mcr_0}-\ref{fig:mcr_1}, respectively.
It can be visually observed from the error maps that our method produces results with fewer residuals compared with the ground truths, which shows that it outperforms other competing methods in correcting color biases and also recovering structural information.

\begin{figure*}[t]
	\setlength{\abovecaptionskip}{0.2cm}
	\centering
	\includegraphics[width=0.95\linewidth]{./sup/mcr_48.pdf}
	\caption{ Visual comparison of competed methods on the MCR dataset of the mono-guided LRIE task. The bottom row refers to the corresponding error maps.}
	\label{fig:mcr_0}
\end{figure*}

\begin{figure*}[t]
	\setlength{\abovecaptionskip}{0.2cm}
	\centering
	\includegraphics[width=0.95\linewidth]{./sup/mcr_105.pdf}
	\caption{ Visual comparison of competed methods on the MCR dataset of the mono-guided LRIE task. The bottom row refers to the corresponding error maps.}
	\label{fig:mcr_1}
\end{figure*}

%% file: girnet_arxiv.bbl
\begin{thebibliography}{60}
\providecommand{\natexlab}[1]{#1}
\providecommand{\url}[1]{\texttt{#1}}
\expandafter\ifx\csname urlstyle\endcsname\relax
  \providecommand{\doi}[1]{doi: #1}\else
  \providecommand{\doi}{doi: \begingroup \urlstyle{rm}\Url}\fi

\bibitem[Backhaus et~al.(2011)Backhaus, Kliegl, and Werner]{backhaus2011color}
Werner~GK Backhaus, Reinhold Kliegl, and John~S Werner.
\newblock \emph{Color vision: Perspectives from different disciplines}.
\newblock Walter de Gruyter, 2011.

\bibitem[Bahdanau et~al.(2014)Bahdanau, Cho, and Bengio]{bahdanau2014neural}
Dzmitry Bahdanau, Kyunghyun Cho, and Yoshua Bengio.
\newblock Neural machine translation by jointly learning to align and translate.
\newblock \emph{arXiv preprint arXiv:1409.0473}, 2014.

\bibitem[Bavirisetti et~al.(2017)Bavirisetti, Kollu, Gang, and Dhuli]{bavirisetti2017medicalfusion}
Durga~Prasad Bavirisetti, Vijayakumar Kollu, Xiao Gang, and Ravindra Dhuli.
\newblock Fusion of mri and ct images using guided image filter and image statistics.
\newblock \emph{International journal of Imaging systems and Technology}, 27\penalty0 (3):\penalty0 227--237, 2017.

\bibitem[Chen et~al.(2018)Chen, Chen, Xu, and Koltun]{chen2018learning}
Chen Chen, Qifeng Chen, Jia Xu, and Vladlen Koltun.
\newblock Learning to see in the dark.
\newblock In \emph{Proceedings of the IEEE conference on computer vision and pattern recognition}, pages 3291--3300, 2018.

\bibitem[Chen et~al.(2021)Chen, Fan, and Panda]{chen2021crossvit}
Chun-Fu~Richard Chen, Quanfu Fan, and Rameswar Panda.
\newblock Crossvit: Cross-attention multi-scale vision transformer for image classification.
\newblock In \emph{Proceedings of the IEEE/CVF international conference on computer vision}, pages 357--366, 2021.

\bibitem[Chen et~al.(2022)Chen, Chu, Zhang, and Sun]{chen2022nafnet}
Liangyu Chen, Xiaojie Chu, Xiangyu Zhang, and Jian Sun.
\newblock Simple baselines for image restoration.
\newblock In \emph{European Conference on Computer Vision}, pages 17--33. Springer, 2022.

\bibitem[Deng et~al.(2022)Deng, Vivone, Paoletti, Scarpa, He, Zhang, Chanussot, and Plaza]{deng2022machine}
Liang-Jian Deng, Gemine Vivone, Mercedes~E Paoletti, Giuseppe Scarpa, Jiang He, Yongjun Zhang, Jocelyn Chanussot, and Antonio Plaza.
\newblock Machine learning in pansharpening: A benchmark, from shallow to deep networks.
\newblock \emph{IEEE Geoscience and Remote Sensing Magazine}, 10\penalty0 (3):\penalty0 279--315, 2022.

\bibitem[Deng and Dragotti(2019)]{deng2019deep}
Xin Deng and Pier~Luigi Dragotti.
\newblock Deep coupled ista network for multi-modal image super-resolution.
\newblock \emph{IEEE Transactions on Image Processing}, 29:\penalty0 1683--1698, 2019.

\bibitem[Deng and Dragotti(2020)]{deng2020cunet}
Xin Deng and Pier~Luigi Dragotti.
\newblock Deep convolutional neural network for multi-modal image restoration and fusion.
\newblock \emph{IEEE transactions on pattern analysis and machine intelligence}, 43\penalty0 (10):\penalty0 3333--3348, 2020.

\bibitem[Dong et~al.(2022)Dong, Xu, Miao, Ma, Zhang, Yang, Jin, Teoh, and Shen]{dong2022abandoning}
Xingbo Dong, Wanyan Xu, Zhihui Miao, Lan Ma, Chao Zhang, Jiewen Yang, Zhe Jin, Andrew Beng~Jin Teoh, and Jiajun Shen.
\newblock Abandoning the bayer-filter to see in the dark.
\newblock In \emph{Proceedings of the IEEE/CVF Conference on Computer Vision and Pattern Recognition}, pages 17431--17440, 2022.

\bibitem[Guo et~al.(2018)Guo, Li, Guo, Cong, Fu, and Han]{guo2018hierarchical}
Chunle Guo, Chongyi Li, Jichang Guo, Runmin Cong, Huazhu Fu, and Ping Han.
\newblock Hierarchical features driven residual learning for depth map super-resolution.
\newblock \emph{IEEE Transactions on Image Processing}, 28\penalty0 (5):\penalty0 2545--2557, 2018.

\bibitem[Hassani and Shi(2022)]{hassani2022dilated}
Ali Hassani and Humphrey Shi.
\newblock Dilated neighborhood attention transformer.
\newblock \emph{arXiv preprint arXiv:2209.15001}, 2022.

\bibitem[Hassani et~al.(2023)Hassani, Walton, Li, Li, and Shi]{hassani2023neighborhood}
Ali Hassani, Steven Walton, Jiachen Li, Shen Li, and Humphrey Shi.
\newblock Neighborhood attention transformer.
\newblock In \emph{Proceedings of the IEEE/CVF Conference on Computer Vision and Pattern Recognition}, pages 6185--6194, 2023.

\bibitem[He et~al.(2010)He, Sun, and Tang]{he2010guided}
Kaiming He, Jian Sun, and Xiaoou Tang.
\newblock Guided image filtering.
\newblock In \emph{European conference on computer vision}, pages 1--14. Springer, 2010.

\bibitem[He et~al.(2012)He, Sun, and Tang]{he2012guided}
Kaiming He, Jian Sun, and Xiaoou Tang.
\newblock Guided image filtering.
\newblock \emph{IEEE transactions on pattern analysis and machine intelligence}, 35\penalty0 (6):\penalty0 1397--1409, 2012.

\bibitem[He et~al.(2021)He, Zhu, Li, Bai, Cong, Zhang, Lin, Liu, and Zhao]{he2021towards}
Lingzhi He, Hongguang Zhu, Feng Li, Huihui Bai, Runmin Cong, Chunjie Zhang, Chunyu Lin, Meiqin Liu, and Yao Zhao.
\newblock Towards fast and accurate real-world depth super-resolution: Benchmark dataset and baseline.
\newblock In \emph{Proceedings of the IEEE/CVF Conference on Computer Vision and Pattern Recognition}, pages 9229--9238, 2021.

\bibitem[Hendrycks and Gimpel(2016)]{hendrycks2016gaussian}
Dan Hendrycks and Kevin Gimpel.
\newblock Gaussian error linear units (gelus).
\newblock \emph{arXiv preprint arXiv:1606.08415}, 2016.

\bibitem[Hirschmuller and Scharstein(2007)]{hirschmuller2007mid}
Heiko Hirschmuller and Daniel Scharstein.
\newblock Evaluation of cost functions for stereo matching.
\newblock In \emph{2007 IEEE conference on computer vision and pattern recognition}, pages 1--8. IEEE, 2007.

\bibitem[Hu et~al.(2023)Hu, Jiang, Liu, and Ma]{hu2023zmff}
Xingyu Hu, Junjun Jiang, Xianming Liu, and Jiayi Ma.
\newblock Zmff: Zero-shot multi-focus image fusion.
\newblock \emph{Information Fusion}, 92:\penalty0 127--138, 2023.

\bibitem[Hui et~al.(2016)Hui, Loy, and Tang]{hui2016depth}
Tak-Wai Hui, Chen~Change Loy, and Xiaoou Tang.
\newblock Depth map super-resolution by deep multi-scale guidance.
\newblock In \emph{Computer Vision--ECCV 2016: 14th European Conference, Amsterdam, The Netherlands, October 11-14, 2016, Proceedings, Part III 14}, pages 353--369. Springer, 2016.

\bibitem[Jin et~al.(2022)Jin, Zhang, Jiang, Vivone, and Deng]{jin2022lagconv}
Zi-Rong Jin, Tian-Jing Zhang, Tai-Xiang Jiang, Gemine Vivone, and Liang-Jian Deng.
\newblock Lagconv: Local-context adaptive convolution kernels with global harmonic bias for pansharpening.
\newblock In \emph{Proceedings of the AAAI Conference on Artificial Intelligence}, pages 1113--1121, 2022.

\bibitem[Khan et~al.(2022)Khan, Naseer, Hayat, Zamir, Khan, and Shah]{khan2022transformers}
Salman Khan, Muzammal Naseer, Munawar Hayat, Syed~Waqas Zamir, Fahad~Shahbaz Khan, and Mubarak Shah.
\newblock Transformers in vision: A survey.
\newblock \emph{ACM computing surveys (CSUR)}, 54\penalty0 (10s):\penalty0 1--41, 2022.

\bibitem[Kim et~al.(2021)Kim, Ponce, and Ham]{kim2021deformable}
Beomjun Kim, Jean Ponce, and Bumsub Ham.
\newblock Deformable kernel networks for joint image filtering.
\newblock \emph{International Journal of Computer Vision}, 129\penalty0 (2):\penalty0 579--600, 2021.

\bibitem[Li et~al.(2016)Li, Huang, Ahuja, and Yang]{li2016djif}
Yijun Li, Jia-Bin Huang, Narendra Ahuja, and Ming-Hsuan Yang.
\newblock Deep joint image filtering.
\newblock In \emph{European conference on computer vision}, pages 154--169. Springer, 2016.

\bibitem[Li et~al.(2019)Li, Huang, Ahuja, and Yang]{li2019joint}
Yijun Li, Jia-Bin Huang, Narendra Ahuja, and Ming-Hsuan Yang.
\newblock Joint image filtering with deep convolutional networks.
\newblock \emph{IEEE transactions on pattern analysis and machine intelligence}, 41\penalty0 (8):\penalty0 1909--1923, 2019.

\bibitem[Li et~al.(2014)Li, Zheng, Zhu, Yao, and Wu]{li2014weighted}
Zhengguo Li, Jinghong Zheng, Zijian Zhu, Wei Yao, and Shiqian Wu.
\newblock Weighted guided image filtering.
\newblock \emph{IEEE Transactions on Image processing}, 24\penalty0 (1):\penalty0 120--129, 2014.

\bibitem[Liang et~al.(2021)Liang, Cao, Sun, Zhang, Van~Gool, and Timofte]{liang2021swinir}
Jingyun Liang, Jiezhang Cao, Guolei Sun, Kai Zhang, Luc Van~Gool, and Radu Timofte.
\newblock Swinir: Image restoration using swin transformer.
\newblock In \emph{Proceedings of the IEEE/CVF international conference on computer vision}, pages 1833--1844, 2021.

\bibitem[Liu et~al.(2017)Liu, Chen, Peng, and Wang]{liu2017multi}
Yu Liu, Xun Chen, Hu Peng, and Zengfu Wang.
\newblock Multi-focus image fusion with a deep convolutional neural network.
\newblock \emph{Information Fusion}, 36:\penalty0 191--207, 2017.

\bibitem[Lu et~al.(2014)Lu, Ren, and Liu]{lu2014depth}
Si Lu, Xiaofeng Ren, and Feng Liu.
\newblock Depth enhancement via low-rank matrix completion.
\newblock In \emph{Proceedings of the IEEE conference on computer vision and pattern recognition}, pages 3390--3397, 2014.

\bibitem[Ma et~al.(2021)Ma, Le, Tian, and Jiang]{ma2021smfuse}
Jiayi Ma, Zhuliang Le, Xin Tian, and Junjun Jiang.
\newblock Smfuse: Multi-focus image fusion via self-supervised mask-optimization.
\newblock \emph{IEEE Transactions on Computational Imaging}, 7:\penalty0 309--320, 2021.

\bibitem[Ma et~al.(2022)Ma, Tang, Fan, Huang, Mei, and Ma]{ma2022swinfusion}
Jiayi Ma, Linfeng Tang, Fan Fan, Jun Huang, Xiaoguang Mei, and Yong Ma.
\newblock Swinfusion: Cross-domain long-range learning for general image fusion via swin transformer.
\newblock \emph{IEEE/CAA Journal of Automatica Sinica}, 9\penalty0 (7):\penalty0 1200--1217, 2022.

\bibitem[Mac~Aodha et~al.(2012)Mac~Aodha, Campbell, Nair, and Brostow]{mac2012patch}
Oisin Mac~Aodha, Neill~DF Campbell, Arun Nair, and Gabriel~J Brostow.
\newblock Patch based synthesis for single depth image super-resolution.
\newblock In \emph{Computer Vision--ECCV 2012: 12th European Conference on Computer Vision, Florence, Italy, October 7-13, 2012, Proceedings, Part III 12}, pages 71--84. Springer, 2012.

\bibitem[Nejati et~al.(2015)Nejati, Samavi, and Shirani]{nejati2015multi}
Mansour Nejati, Shadrokh Samavi, and Shahram Shirani.
\newblock Multi-focus image fusion using dictionary-based sparse representation.
\newblock \emph{Information Fusion}, 25:\penalty0 72--84, 2015.

\bibitem[Pan et~al.(2019)Pan, Dong, Ren, Lin, Tang, and Yang]{pan2019spatially}
Jinshan Pan, Jiangxin Dong, Jimmy~S Ren, Liang Lin, Jinhui Tang, and Ming-Hsuan Yang.
\newblock Spatially variant linear representation models for joint filtering.
\newblock In \emph{Proceedings of the IEEE/CVF Conference on Computer Vision and Pattern Recognition}, pages 1702--1711, 2019.

\bibitem[Ronneberger et~al.(2015)Ronneberger, Fischer, and Brox]{ronneberger2015u}
Olaf Ronneberger, Philipp Fischer, and Thomas Brox.
\newblock U-net: Convolutional networks for biomedical image segmentation.
\newblock In \emph{International Conference on Medical image computing and computer-assisted intervention}, pages 234--241. Springer, 2015.

\bibitem[{Shutao Li} et~al.(2013){Shutao Li}, {Xudong Kang}, and {Jianwen Hu}]{ShutaoLiImageFusion2013}
{Shutao Li}, {Xudong Kang}, and {Jianwen Hu}.
\newblock Image {{Fusion With Guided Filtering}}.
\newblock \emph{IEEE Transactions on Image Processing}, 22\penalty0 (7):\penalty0 2864--2875, 2013.

\bibitem[Silberman et~al.(2012)Silberman, Hoiem, Kohli, and Fergus]{silberman2012indoor}
Nathan Silberman, Derek Hoiem, Pushmeet Kohli, and Rob Fergus.
\newblock Indoor segmentation and support inference from rgbd images.
\newblock In \emph{Computer Vision--ECCV 2012: 12th European Conference on Computer Vision, Florence, Italy, October 7-13, 2012, Proceedings, Part V 12}, pages 746--760. Springer, 2012.

\bibitem[Su et~al.(2019)Su, Jampani, Sun, Gallo, Learned-Miller, and Kautz]{Su_2019_pac}
Hang Su, Varun Jampani, Deqing Sun, Orazio Gallo, Erik Learned-Miller, and Jan Kautz.
\newblock Pixel-adaptive convolutional neural networks.
\newblock In \emph{Proceedings of the IEEE/CVF Conference on Computer Vision and Pattern Recognition (CVPR)}, 2019.

\bibitem[Tan and Bansal(2019)]{tan2019lxmert}
Hao Tan and Mohit Bansal.
\newblock Lxmert: Learning cross-modality encoder representations from transformers.
\newblock In \emph{Proceedings of the 2019 Conference on Empirical Methods in Natural Language Processing and the 9th International Joint Conference on Natural Language Processing (EMNLP-IJCNLP)}, pages 5100--5111, 2019.

\bibitem[Tomasi and Manduchi(1998)]{tomasi1998bilateral}
Carlo Tomasi and Roberto Manduchi.
\newblock Bilateral filtering for gray and color images.
\newblock In \emph{Sixth international conference on computer vision (IEEE Cat. No. 98CH36271)}, pages 839--846. IEEE, 1998.

\bibitem[Vaswani et~al.(2017)Vaswani, Shazeer, Parmar, Uszkoreit, Jones, Gomez, Kaiser, and Polosukhin]{vaswani2017attention}
Ashish Vaswani, Noam Shazeer, Niki Parmar, Jakob Uszkoreit, Llion Jones, Aidan~N Gomez, {\L}ukasz Kaiser, and Illia Polosukhin.
\newblock Attention is all you need.
\newblock \emph{Advances in neural information processing systems}, 30, 2017.

\bibitem[Vivone et~al.(2014)Vivone, Alparone, Chanussot, Dalla~Mura, Garzelli, Licciardi, Restaino, and Wald]{vivone2014critical}
Gemine Vivone, Luciano Alparone, Jocelyn Chanussot, Mauro Dalla~Mura, Andrea Garzelli, Giorgio~A Licciardi, Rocco Restaino, and Lucien Wald.
\newblock A critical comparison among pansharpening algorithms.
\newblock \emph{IEEE Transactions on Geoscience and Remote Sensing}, 53\penalty0 (5):\penalty0 2565--2586, 2014.

\bibitem[Wang et~al.(2018)Wang, Fang, Li, and Zhang]{wang2018bayesian}
Tingting Wang, Faming Fang, Fang Li, and Guixu Zhang.
\newblock High-quality bayesian pansharpening.
\newblock \emph{IEEE Transactions on Image Processing}, 28\penalty0 (1):\penalty0 227--239, 2018.

\bibitem[Wang et~al.(2004)Wang, Bovik, Sheikh, and Simoncelli]{wang2004image}
Zhou Wang, Alan~C Bovik, Hamid~R Sheikh, and Eero~P Simoncelli.
\newblock Image quality assessment: from error visibility to structural similarity.
\newblock \emph{IEEE transactions on image processing}, 13\penalty0 (4):\penalty0 600--612, 2004.

\bibitem[Wu et~al.(2018)Wu, Zheng, Zhang, and Huang]{wu2018fgf}
Huikai Wu, Shuai Zheng, Junge Zhang, and Kaiqi Huang.
\newblock Fast end-to-end trainable guided filter.
\newblock In \emph{Proceedings of the IEEE Conference on Computer Vision and Pattern Recognition}, pages 1838--1847, 2018.

\bibitem[Xiao and Gan(2012)]{xiao2012fast}
Chunxia Xiao and Jiajia Gan.
\newblock Fast image dehazing using guided joint bilateral filter.
\newblock \emph{The Visual Computer}, 28:\penalty0 713--721, 2012.

\bibitem[Xiong et~al.(2021)Xiong, Wang, Heidrich, and Nayar]{xiong2021mffnet}
Jinhui Xiong, Jian Wang, Wolfgang Heidrich, and Shree Nayar.
\newblock Seeing in extra darkness using a deep-red flash.
\newblock In \emph{Proceedings of the IEEE/CVF Conference on Computer Vision and Pattern Recognition}, pages 10000--10009, 2021.

\bibitem[Xu et~al.(2020)Xu, Ma, Le, Jiang, and Guo]{xu2020fusiondn}
Han Xu, Jiayi Ma, Zhuliang Le, Junjun Jiang, and Xiaojie Guo.
\newblock Fusiondn: A unified densely connected network for image fusion.
\newblock In \emph{Proceedings of the AAAI conference on artificial intelligence}, pages 12484--12491, 2020.

\bibitem[Xu et~al.(2021)Xu, Zhang, Zhao, Sun, Liu, and Zhang]{xu2021gppnn}
Shuang Xu, Jiangshe Zhang, Zixiang Zhao, Kai Sun, Junmin Liu, and Chunxia Zhang.
\newblock Deep gradient projection networks for pan-sharpening.
\newblock In \emph{Proceedings of the IEEE/CVF Conference on Computer Vision and Pattern Recognition}, pages 1366--1375, 2021.

\bibitem[Yang et~al.(2017)Yang, Fu, Hu, Huang, Ding, and Paisley]{yang2017pannet}
Junfeng Yang, Xueyang Fu, Yuwen Hu, Yue Huang, Xinghao Ding, and John Paisley.
\newblock Pannet: A deep network architecture for pan-sharpening.
\newblock In \emph{Proceedings of the IEEE international conference on computer vision}, pages 5449--5457, 2017.

\bibitem[Yuhas et~al.(1992)Yuhas, Goetz, and Boardman]{yuhas1992discrimination}
Roberta~H Yuhas, Alexander~FH Goetz, and Joe~W Boardman.
\newblock Discrimination among semi-arid landscape endmembers using the spectral angle mapper (sam) algorithm.
\newblock In \emph{JPL, Summaries of the Third Annual JPL Airborne Geoscience Workshop. Volume 1: AVIRIS Workshop}, 1992.

\bibitem[Zhang et~al.(2020)Zhang, Liao, Liu, Ma, Yang, and Xue]{zhang2020real}
Juncheng Zhang, Qingmin Liao, Shaojun Liu, Haoyu Ma, Wenming Yang, and Jing-Hao Xue.
\newblock Real-mff: A large realistic multi-focus image dataset with ground truth.
\newblock \emph{Pattern Recognition Letters}, 138:\penalty0 370--377, 2020.

\bibitem[Zhang et~al.(2018)Zhang, Isola, Efros, Shechtman, and Wang]{zhang2018lpips}
Richard Zhang, Phillip Isola, Alexei~A Efros, Eli Shechtman, and Oliver Wang.
\newblock The unreasonable effectiveness of deep features as a perceptual metric.
\newblock In \emph{Proceedings of the IEEE conference on computer vision and pattern recognition}, pages 586--595, 2018.

\bibitem[Zhang et~al.(2022)Zhang, Zhao, Zhang, Peng, and Fan]{zhang2022segmentation}
Xiang Zhang, Wanqing Zhao, Wei Zhang, Jinye Peng, and Jianping Fan.
\newblock Guided filter network for semantic image segmentation.
\newblock \emph{IEEE Transactions on Image Processing}, 31:\penalty0 2695--2709, 2022.

\bibitem[Zhao et~al.(2022)Zhao, Zhang, Xu, Lin, and Pfister]{zhao2022dct}
Zixiang Zhao, Jiangshe Zhang, Shuang Xu, Zudi Lin, and Hanspeter Pfister.
\newblock Discrete cosine transform network for guided depth map super-resolution.
\newblock In \emph{Proceedings of the IEEE/CVF Conference on Computer Vision and Pattern Recognition}, pages 5697--5707, 2022.

\bibitem[Zhao et~al.(2023{\natexlab{a}})Zhao, Bai, Zhang, Zhang, Xu, Lin, Timofte, and Van~Gool]{zhao2023cddfuse}
Zixiang Zhao, Haowen Bai, Jiangshe Zhang, Yulun Zhang, Shuang Xu, Zudi Lin, Radu Timofte, and Luc Van~Gool.
\newblock Cddfuse: Correlation-driven dual-branch feature decomposition for multi-modality image fusion.
\newblock In \emph{Proceedings of the IEEE/CVF Conference on Computer Vision and Pattern Recognition}, pages 5906--5916, 2023{\natexlab{a}}.

\bibitem[Zhao et~al.(2023{\natexlab{b}})Zhao, Zhang, Gu, Tan, Xu, Zhang, Timofte, and Van~Gool]{zhao2023spherical}
Zixiang Zhao, Jiangshe Zhang, Xiang Gu, Chengli Tan, Shuang Xu, Yulun Zhang, Radu Timofte, and Luc Van~Gool.
\newblock Spherical space feature decomposition for guided depth map super-resolution.
\newblock In \emph{Proceedings of the IEEE/CVF International Conference on Computer Vision (ICCV)}, pages 12547--12558, 2023{\natexlab{b}}.

\bibitem[Zhong et~al.(2021{\natexlab{a}})Zhong, Liu, Jiang, Zhao, Chen, and Ji]{zhong2021high}
Zhiwei Zhong, Xianming Liu, Junjun Jiang, Debin Zhao, Zhiwen Chen, and Xiangyang Ji.
\newblock High-resolution depth maps imaging via attention-based hierarchical multi-modal fusion.
\newblock \emph{IEEE Transactions on Image Processing}, 31:\penalty0 648--663, 2021{\natexlab{a}}.

\bibitem[Zhong et~al.(2021{\natexlab{b}})Zhong, Liu, Jiang, Zhao, and Ji]{zhong2021dagif}
Zhiwei Zhong, Xianming Liu, Junjun Jiang, Debin Zhao, and Xiangyang Ji.
\newblock Deep attentional guided image filtering.
\newblock \emph{arXiv preprint arXiv:2112.06401}, 2021{\natexlab{b}}.

\bibitem[Zhou et~al.(2023)Zhou, Yan, Pan, Ren, Xie, and Cao]{madunet}
Man Zhou, Keyu Yan, Jinshan Pan, Wenqi Ren, Qi Xie, and Xiangyong Cao.
\newblock Memory-augmented deep unfolding network for guided image super-resolution.
\newblock \emph{International Journal of Computer Vision}, 131\penalty0 (1):\penalty0 215--242, 2023.

\end{thebibliography}
